\documentclass[11pt,letterpaper]{article}
\usepackage{mmpreprint}

\title{General progress substitutes for\\specific training data in language models}
\author{Mark Marosi\affil{1} \and Peter Antal\affil{1}}
\affiliation{1}{Budapest University of Technology and Economics (BME), Budapest, Hungary}
\date{\today}

\begin{document}

\maketitle

\begin{abstract}
Language models acquire their capabilities from training data, and developers withhold data during training to keep models from learning capabilities that could be misused. How long this delays a capability, and how much data is needed to acquire it later, is unknown. Here we show that the general progress a model makes on the rest of its training data substitutes for the data that was withheld. We withheld the training signal for individual skills in models of chess, Othello, molecular strings, bracket languages and web text. The delay of a withheld skill in larger models was predicted from smaller ones. A chess model never explicitly trained to respond to check ranked the legal responses almost as well as a model that was. When models were then trained on the withheld data, they acquired the skill with a small fraction of the data needed from initialization, and this fraction decreased as the models progressed. A control rule of equal difficulty was learned much more slowly. A programming language withheld from the training text of a 7-billion-parameter model was regained in the same way. Data filtering therefore delays a capability, and the data required to regain it decreases as models improve.
\end{abstract}

\section{Introduction}

Developers shape the capabilities of language models through their training data. Mixtures are tuned to balance domains~\citep{ref23,ref24,ref25}, and documents or individual targets are removed to withhold a capability that could be misused~\citep{ref2,ref26,ref3}. The effect of such a filter depends on how long it holds the capability back as the model improves and on how much data the model needs to acquire the capability once the data return. Scaling laws relate compute, data and model size to loss~\citep{ref5,ref6}, and loss indexes when skills appear~\citep{ref4,ref44}. Neither gives how much of a skill comes from its own examples and how much from the rest of the training data.

Earlier results show that the rest of the corpus contributes to a skill. Pretraining reduces the data a later task needs~\citep{hernandez2021scaling,tripuraneni2020theory}, capabilities suppressed by pruning or unlearning are regained with little further training~\citep{lo2024large,deeb2024unlearning}, and progress accumulates before a score moves~\citep{barak2022hidden,nanda2023progress}. The closest study is that of Rathi and Radford~\citep{ref26}, who filtered a medical domain from models of 61M to 1.8B parameters. They found that filtering becomes more effective with scale and that adversarial fine-tuning restores the unfiltered baseline with less than 0.2\% of the pretraining tokens, a share that falls with model size. These studies do not give the size of the delay, how the cost of acquiring a withheld capability depends on the model's progress, or a control for whether the saving is specific to the withheld skill.

Here we show that withholding a capability's training data delays the capability by a predictable amount, that progress on the rest of the corpus substitutes for the withheld data, and that acquiring the capability once its data return becomes cheaper as models improve. We use target thinning, which removes a fixed fraction of a skill's next-token targets from the training loss while every input token stays in the text, and then return the targets. Models are compared at matched background loss, the prediction error on held-out text outside the thinned targets. The experiments cover chess and Othello~\citep{ref28,ref29,ref30,ref31,mcgrath2022alphazero}, molecular SMILES strings~\citep{weininger1988smiles,schwaller2019molecular,ref32}, Dyck languages~\citep{hewitt2020dyck,ref33} and web text, in models of up to 7 billion parameters.

For king safety in chess, and for the capitals of countries in a 7-billion-parameter web-text model, the delay follows, with no fitted parameter, from a model that keeps all of the skill's targets and one that keeps none; for the other skills the same form holds with fitted parameters. A model that never received the targets has learned much of the skill, and when the targets return it acquires the skill with a small fraction of the data needed from initialization. This cost falls as the model progresses: for each logit by which a supervised model at the same progress exceeds the acquisition level, about tenfold in chess and web text and fourfold in SMILES. These relations predict both trends reported by Rathi and Radford. The deficit that a filter leaves widens with progress, and the same progress makes the capability cheaper to acquire once its data return.

\section{Results}

\subsection{Delay from thinning a skill's targets}

We asked by how much a skill is delayed when only a fraction of its examples is supervised. We trained decoder-only transformers whose hyperparameters transfer across scales~\citep{ref38,ref39} and kept a fraction $f$ of the next-token targets selected by a skill-specific rule, the retained fraction, while every input token stayed in the sequence (Fig.~1a). In chess the rule is king safety, which applies when the side to move is in check, on about 7\% of boards. King-safety competence $q$ is the probability that the model assigns to legal escapes from check, relative to legal plus king-unsafe moves. Background loss $L$ is the held-out prediction error on all positions outside the thinned targets. Comparing models at the same $L$ separates the effect of a skill's own targets from progress on the rest of the corpus (Fig.~B1).

We measured the delay as a shift of onset, the background loss at which competence reaches a fixed level. In 9.7M-parameter chess models, keeping 10\% of in-check targets moved onset at a competence of 0.70 by 0.40 nats towards lower background loss (Fig.~1e). Removing an equal density of random targets left onset unchanged, and giving every in-check target a loss weight of 0.1 produced the same competence as keeping 10\% at full weight (Fig.~B2h,i).

\begin{figure}[p]
\ppfigure{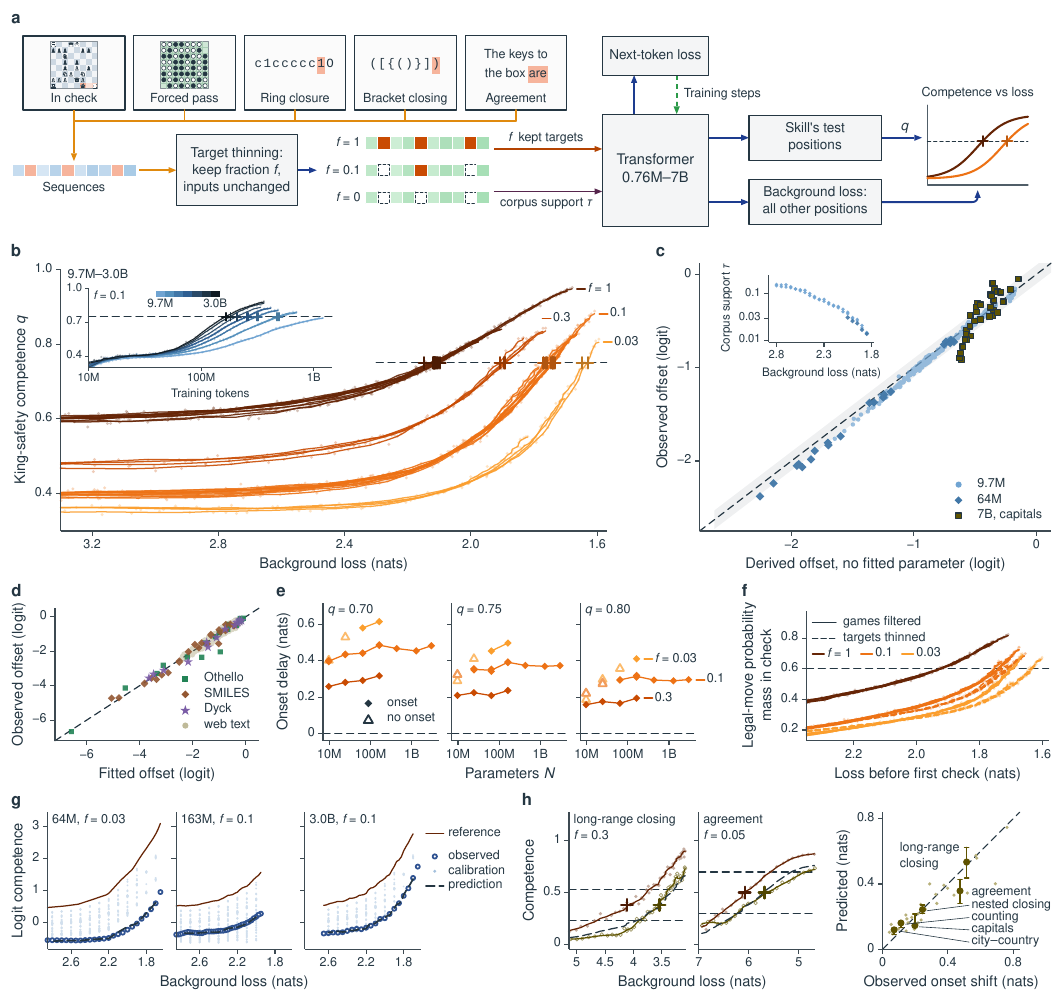}{fig1}{Fig.~1}{Withholding a skill's targets delays it by a predictable amount.}{\textbf{a}, Target thinning keeps a fraction $f$ of a skill's targets; every input stays in the sequence. Competence $q$ is read on the skill's test positions, background loss on all others. \textbf{b}, Chess king-safety competence against background loss for 30 runs at 9.7M--3.0B parameters (colour, $f$; dots, snapshots; lines, five-snapshot smoothing in logit; $+$, onset at $q=0.75$). Inset, the 11 runs with $f=0.1$ against training tokens (colour, model size). \textbf{c}, Observed against derived log-odds offsets from the fully supervised model, with no fitted parameter; one point per thinned model and background loss: 15 chess models at 9.7M (circles; two seeds, $f=0.05$--0.7; 300 points), four at 64M (diamonds; $f=0.01$--0.3; 20 points) and 7B web-text models with $f=0.3$ for capitals (squares; two seeds, 32 points). Shading, $\pm0.1$ logit; dashed, equality. Inset, corpus support $\tau$ against background loss (chess models). \textbf{d}, Observed against fitted offsets for Othello (4 skills), SMILES (6) and Dyck languages (3) at 9.7M, and 770 web-text agreement offsets (34 thinned runs). \textbf{e}, Onset delay relative to the fully supervised model of the same size at three competence levels for the 19 thinned runs of \textbf{b} (lines, size means; triangles, lower bounds). \textbf{f}, Legal-move probability mass in check for 30 runs at 9.7M--64M: fully supervised, games containing a check filtered to a residual share of 0.1 or 0.03 (solid), and in-check targets thinned to the same shares (dashed); colour, share. \textbf{g}, Chess predictions for the median-error trajectory at a fraction and a size omitted from calibration, and for 3.0B calibrated up to 431M. \textbf{h}, Two skills at 7B predicted from the 60M--1B response ($+$, onsets at competence 0.5), and predicted against observed onset shift for six skills (large points, means over five competence levels; bars, range over calibration sizes).}
\end{figure}

For king safety in chess, the delay can be derived from a model that keeps all of the targets and one that keeps none. If the prediction at a test position combines what the skill's own targets teach, with weight $f$, and what the rest of the corpus teaches, the odds of a legal escape are linear in the retained fraction,
\begin{equation}
\operatorname{odds}_x(L,f)=f\,\operatorname{odds}_x(L,1)+(1-f)\,\operatorname{odds}_x(L,0).
\label{eq:odds}
\end{equation}
The terms on the right are the odds at position $x$ of a fully supervised and a never-supervised model at the same background loss. With competence expressed as its log odds, thinning lowers the log odds at every position by
\begin{equation}
\operatorname{logit}q_x(L,f)-\operatorname{logit}q_x(L,1)=\ln\!\left(\frac{f+\tau_x}{1+\tau_x}\right),\qquad \frac{\tau_x}{1+\tau_x}=\frac{\operatorname{odds}_x(L,0)}{\operatorname{odds}_x(L,1)}.
\label{eq:effective-supervision}
\end{equation}
Here $q_x$ is the competence at position $x$ and $\tau_x$ is the corpus support, the contribution of the rest of the corpus in units of what the skill's own targets add. The competence $q(L,f)$ pools the positions. We use one corpus support $\tau(L)$ for all positions, read from the never-supervised model as the value at which $f=0$ reproduces its pooled competence. The ratio $(f+\tau)/(1+\tau)$ is the effective supervision.

We tested the relation on thinned chess models. With the shift applied to every position of the fully supervised model and corpus support read from the never-supervised model, it gave the offsets of 15 thinned 9.7M models with a root-mean-square error of 0.039 logits, and those of 64M models at retained fractions down to 0.01 with an error of 0.089 logits (Fig.~1c). Competence $q$ pools the probability mass of all test positions, and pooling scales the response by a sensitivity of 0.55--0.67. With this sensitivity, the pooled form describes 4,312 offsets from 109 chess trajectories at retained fractions from 0 to 0.7 with a residual standard deviation of 0.034 logits (Fig.~B2k). The relation also gave the offsets of 7-billion-parameter web-text models that kept 30\% of the targets stating a country's capital. In each of two seeds, the offsets derived from a fully supervised and a never-supervised model of that seed differed from the measured ones by 0.14 logits (root mean square) at 16 background losses, less than half the 0.31 logits by which the measured offsets of the two seeds differed; the retained fraction alone missed them by 0.61 logits (Fig.~1c and Fig.~B2n). Othello, SMILES, Dyck languages and web text show the delay (Fig.~B4) and follow the pooled form with their own fitted sensitivities (Fig.~1d). With no fitted parameter, the relation missed the offsets of five skills of 9.7M models in these domains by 0.26--0.80 logits and those of a Dyck skill by 1.7 logits; fitted at each background loss, it described all six with errors of 0.10--0.42 logits.

Corpus support falls as a model improves, from 0.15--0.16 at 2.8 nats of background loss to 0.014--0.015 at 1.85 nats, and it shows no trend with model size. The deficit of a model that lacks a share of the targets therefore widens with progress. Between 2.6 and 1.8 nats the offset grew by 0.22 logits for models that kept 30\% of the targets and by 1.39 logits for models that kept none (Fig.~B2l,m).

The delay in background loss changed little with model size. From 9.7M to 3.0B parameters under a matched schedule, keeping 10\% of in-check targets delayed onset by 0.29--0.49 nats at competence levels $q\geq0.70$, and the delay changed by 0.00--0.03 nats for each tenfold increase in size (Fig.~1e and Fig.~B3). Filtering whole documents had a similar effect. Removing chess games that contain a check produced 78--87\% of the delay obtained by thinning in-check targets to the share that the filter left (Fig.~1f).

Because the response to thinning is stable across sizes, we calibrated it on small models and applied it to larger ones. Each prediction adds this response to the larger model's own fully supervised trajectory, with no parameter fitted to the thinned model. For chess fractions, sizes and learning-rate schedules~\citep{hagele2024scaling} omitted from calibration, median errors were 0.026--0.040 logits, and the response calibrated up to 431M parameters predicted models of 1.28B and 3.0B (Fig.~1g and Fig.~B2a,b). On web text, responses calibrated at 60M--1B predicted the onset shifts of six thinned skills in 7-billion-parameter models with a mean error of 0.058 nats (Fig.~1h and Fig.~B5).

\subsection{Outputs of a never-supervised model}

The derivation attributes part of a skill to the rest of the corpus, so we examined what a model trained without a skill's targets predicts at the skill's positions. We assigned every output token at 8,313 held-out in-check positions to a class of move (Fig.~2a). A chess model trained without in-check targets placed about half of its probability on moves that would be legal if the king were not in check, 0.48 for the 64M model at its last snapshot against 0.02 for the fully supervised model. Where two or more replies were legal, its most probable legal reply was the move played in the game at 0.64 of positions, against 0.68 for the fully supervised model and 0.30 by chance (Fig.~2b).

Under target thinning, replies to check remain in the text the model reads. We therefore trained chess models on games in which no check occurs. At the first check of a held-out game, these models ranked the move played first among the legal replies in 0.65--0.68 of games at 9.7M--163M parameters, against 0.68--0.71 for fully supervised models. Their preference followed the position: in pairs of positions in which the same two quiet moves exchange legality, they preferred the legal move on both boards in 0.65 of pairs at 9.7M and 0.78 at 163M, against 0.93--0.97 for fully supervised models and zero for a ranking that ignores the board (Fig.~2e--g).

\begin{figure}[p]
\ppfigure{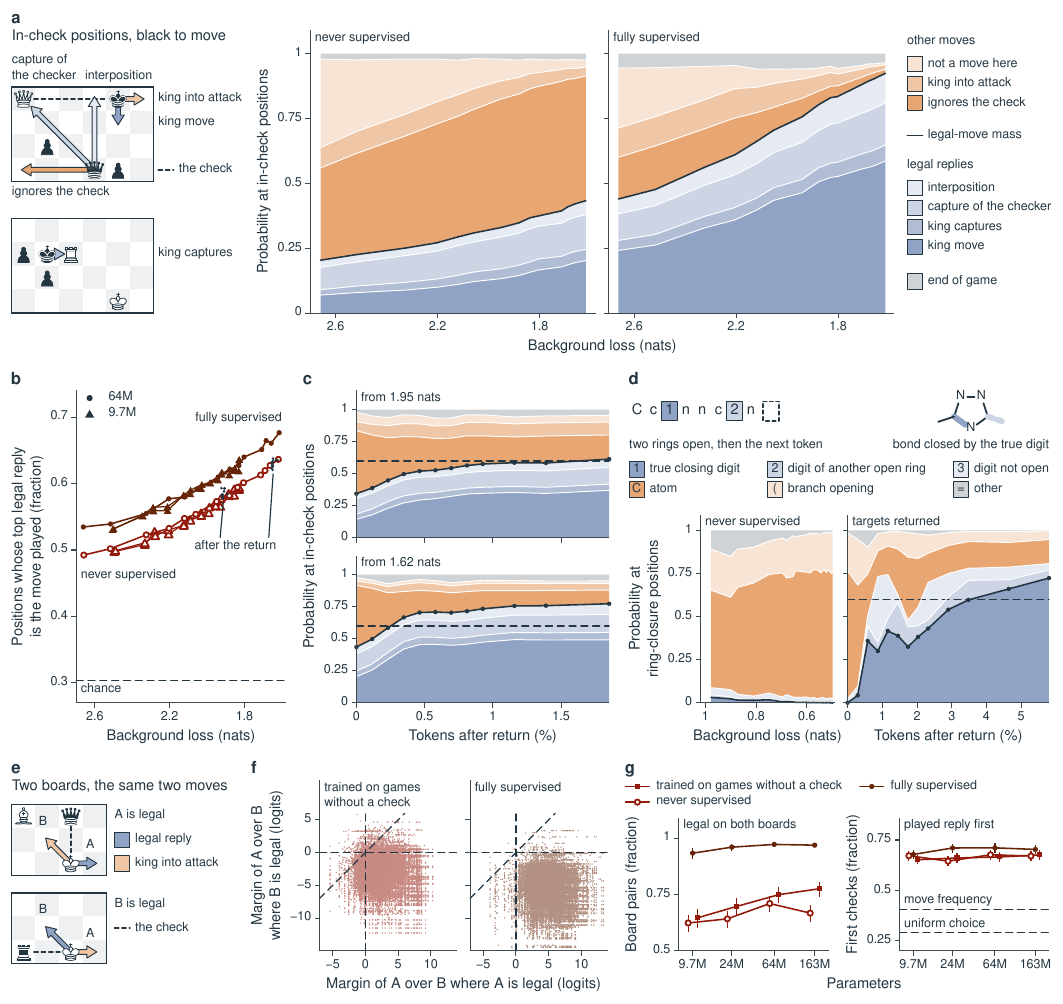}{fig2}{Fig.~2}{What a never-supervised model does at the skill's positions, and what returned targets add.}{\textbf{a}, Next-move probability at 8,313 held-out in-check positions, stacked by class of move, against background loss, for 64M chess models trained without in-check targets (left) and fully supervised (right); line, legal-move mass. Key, two of these positions, one arrow per class; not a move here, tokens no piece on the board can play. \textbf{b}, Fraction of in-check positions with $\geq2$ legal replies whose most probable legal reply is the move played, at 64M and 9.7M (two seeds); dots, the saved states of \textbf{c}; dashed, chance. \textbf{c}, The classes of \textbf{a} after all in-check targets return to snapshots of the never-supervised 64M model at 1.95 and 1.62 nats, against tokens after the return as a percentage of the tokens the 64M model needs from initialization to reach a legal-move mass of 0.6 (dashed); dots, saved states after 0--16 steps. \textbf{d}, SMILES, 9.7M model trained without ring-closure targets: next-token probability at 23,190 ring-closure positions by class of token (line, true closing digit), against background loss (left) and after all ring-closure targets return to the last snapshot (right; tokens as a percentage of those from initialization; dashed, 0.6; dots, saved states). \textbf{e}, Two held-out first checks on which the same two king moves, A and B, exchange legality (colours as in \textbf{a}). \textbf{f}, Logit margin of A over B where A is legal against that where B is legal, for 18,701 board pairs at 64M; dashed, zero and equal margins. \textbf{g}, Board pairs with the legal move preferred on both boards, and first checks with the move played ranked first, by model size (95\% intervals; dashed, baselines). Probabilities are means over positions; one run per model and return.}
\end{figure}

When the in-check targets were returned to the training loss of the 64M model, the first million tokens, 0.9\% of the tokens needed from initialization, moved a third of the probability away from moves that ignore the check, most of it to moves of the king (Fig.~2c). The ranking among the legal replies did not change (Fig.~2b). The gain over these first steps was larger at lower background loss, 0.30 against 0.23 of the probability.

Supervision on one kind of reply did not transfer to another. From the 9.7M model trained on games without a check, 3,072 supervised replies to knight checks, which cannot be blocked, raised legal-move mass at held-out knight checks from 0.57 to 0.90 and lowered it from 0.39 to 0.35 where every legal reply is a block, as did the same number of ordinary moves; supervised blocks raised it to 0.89 there (Fig.~C1).

Molecular strings showed the same sequence from a lower starting point. Without ring-closure targets, a SMILES model wrote a further atom where a ring should close and gave the closing digit a probability of 0.0004. Two optimizer steps after the targets returned, the digit had a probability of 0.36 (Fig.~2d).

\subsection{Cost of acquiring a withheld skill}

We measured how much data a never-supervised model needs to acquire the skill once its targets return. We returned the targets at checkpoints along the training of a source model, a model trained without the skill's targets, and trained until competence reached a fixed level, the acquisition level (Fig.~3a). The acquisition cost is the number of tokens this took, relative to the tokens a same-size model needs to reach the level from initialization~\citep{hernandez2021scaling}. When the targets returned, chess source models had a king-safety competence of 0.32--0.50, below the acquisition level. From each source run's lowest-loss checkpoint, acquisition required at most 0.10--0.78\% of the tokens needed from initialization, and 3.1--7.5\% when the 50 preceding steps that restore the optimizer state, which contain none of the returned targets, are also counted (Fig.~D1e).

To test whether this saving is specific to the withheld skill, we used a control rule that applies one fixed permutation of the move labels to every in-check target. The permuted rule matches the true escapes in entropy and attainable accuracy, but its targets are the permuted labels and not the escaping moves, and from initialization the two rules needed the same number of tokens to within a factor of 1.06. From the same source checkpoints, the permuted rule needed 2.4--16 times the tokens of the true rule (Fig.~3b). The cost changed little with batch size. For returns of 25 steps or more, a fourfold smaller batch needed 1.1--1.65 times the tokens in 4.3--6.6 times as many steps (Fig.~D1j).

\begin{figure}[p]
\ppfigure{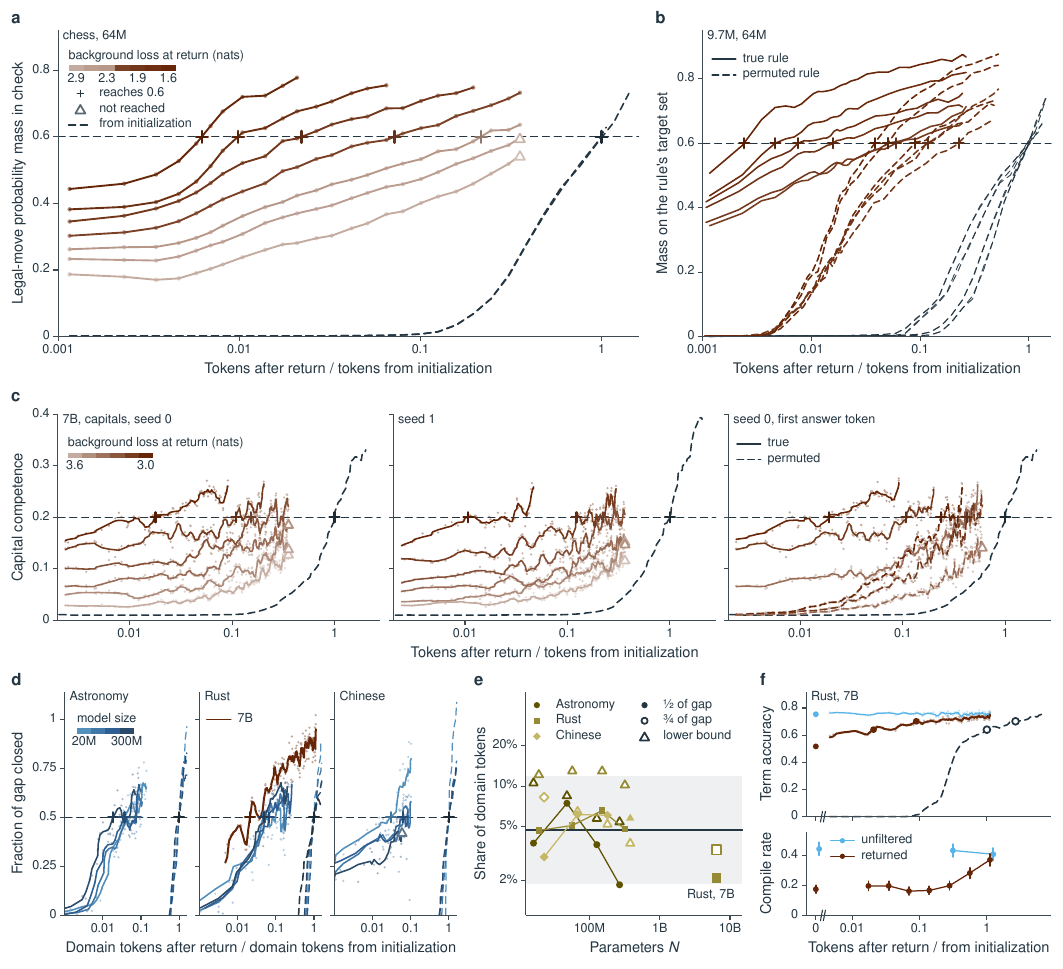}{fig3}{Fig.~3}{Withheld skills return at a fraction of their cost, and the saving is specific.}{Token axes: tokens after the return over the same-size model's tokens from initialization to the same level (dashed curves, from initialization). \textbf{a}, Legal-move mass in check after in-check targets return to a 64M chess model, at seven background losses (colour; dots, evaluations; $+$, mass 0.6 reached; triangles, not reached). \textbf{b}, Mass on each rule's target set after the true (solid) or a permuted (dashed) rule returns, six checkpoints at 9.7M and 64M; each rule relative to its own tokens from initialization. \textbf{c}, Capital competence of 7B models after capital targets return at six checkpoints, seeds 0 and 1 (colour, background loss at return; lines, five-evaluation smoothing in logit; $+$, first crossing of 0.20), and the true against the permuted first answer token from three checkpoints of seed 0 (right). \textbf{d}, Fraction of the term-accuracy gap closed after the unfiltered stream returns to models trained without astronomy, Rust or Chinese documents (colour, model size), against domain tokens relative to the unfiltered model's at half the gap; $+$, half closed; heavy line, Rust at 7B (filtered to 3\%). \textbf{e}, Share of domain tokens needed to close half (filled) or three quarters (open) of the gap (triangles, lower bounds); line and band, constant fitted at 20M--300M and its 95\% prediction interval; large squares, Rust at 7B. \textbf{f}, Rust at 7B: term accuracy (top; circles, half and three quarters of the gap closed; open, from initialization) and compile rate of 2,120 sampled completions of 212 function tasks (bottom; bars, 95\% bootstrap intervals over tasks) for the returned (brown) and the continued unfiltered model (blue); left of the break, at the return. One run per checkpoint, rule and seed (\textbf{a}--\textbf{c}); 12 continuation runs (\textbf{d}, \textbf{e}); one run per model (\textbf{f}).}
\end{figure}

We asked what determines the cost across checkpoints, model sizes and returned shares. For a returned share $s$, the acquisition loss is the background loss at which a model trained with that share from the start reaches the acquisition level. The logit distance is the log odds of the acquisition level minus the log odds of a same-size model trained with the returned share, read at the background loss at which the targets return. A negative distance, which we call a surplus, means that the source model has passed the acquisition loss. Source models that received their targets before reaching the acquisition loss acquired the skill when their own background loss reached it, with a median difference of at most 0.04 nats, and their cost was the training needed to get there (Fig.~4b and Fig.~D2h).

Below the acquisition loss, each logit of surplus lowered the cost about tenfold in chess and web text and fourfold in SMILES. The cost fell by $1.12\pm0.04$ decades per logit in chess, $0.91\pm0.04$ in web text and $0.59\pm0.07$ in SMILES, in 312 runs from 27 source runs (Fig.~4a,c and Fig.~D2). The rates differed between the domains. At its acquisition loss a model needed 30\%, 24\% and 38\% of the tokens from initialization in chess, web text and SMILES. In Othello, never-supervised models acquired the forced-pass rule within about eight optimizer steps at every distance up to 0.85 logits above the acquisition loss (Fig.~D2f).

Logit distance predicted the cost better than the model's score. Among 207 runs from source models scoring near their floor, logit distance ordered 85\% of the pairs within a domain by their cost, whereas the score when the targets returned ordered 53\% (Fig.~4d).

\begin{figure}[t]
\ppfigure{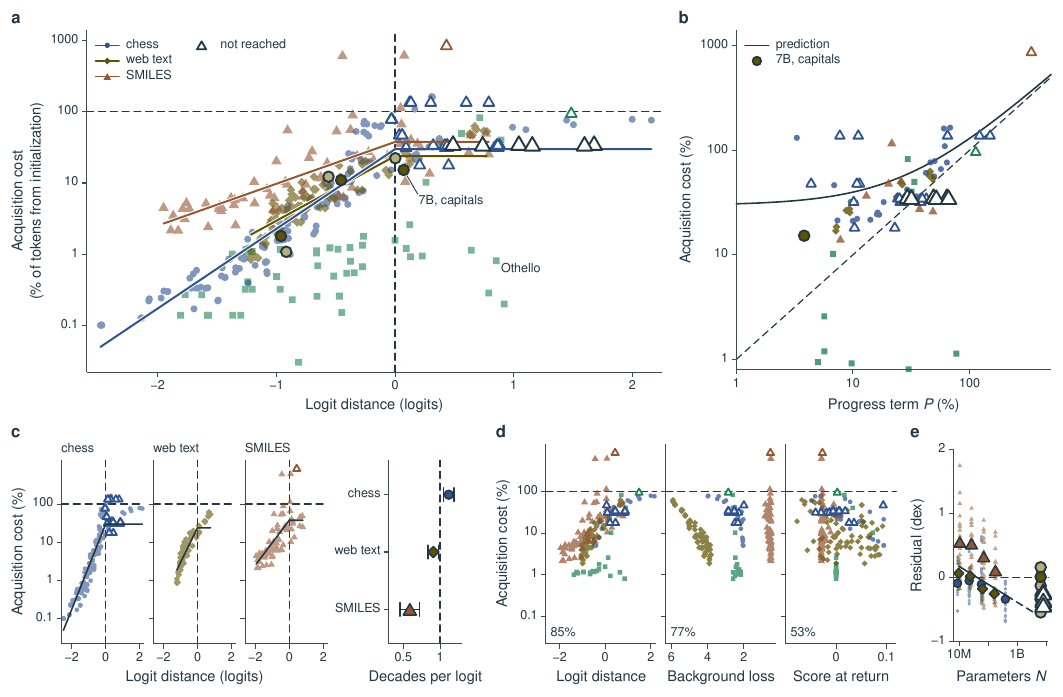}{fig4}{Fig.~4}{Acquisition cost falls about tenfold per logit of surplus in chess and web text and fourfold in SMILES.}{\textbf{a}, Acquisition cost (tokens to the acquisition level after the targets return, per cent of same-size tokens from initialization) against logit distance at the return (dashed vertical, acquisition loss). Runs at 9.7M--431M (chess 186, web text 88, SMILES 66, Othello 50); open triangles, lower bounds. Lines, the two-part relation with an intercept and a rate for each domain, fitted jointly with a random intercept per source run (312 runs, 27 source runs) and drawn over each domain's range; Othello is not fitted. Large markers, capitals at 7B (dark, seed 0; light, seed 1), not fitted. \textbf{b}, Runs returned above the acquisition loss: cost against the progress term $P$ (97 of 108 runs, seven at 7B; 11 with $P<1\%$ not drawn). Line, $P+10^{-0.53}$, from the single least-squares line through the 312 runs, $10^{-0.53+0.97x}$; dashed, cost $=P$. \textbf{c}, Each fitted domain with its line of \textbf{a}, and the three rates (bars, 95\% intervals; dashed, one decade per logit). \textbf{d}, The 207 runs from sources scoring near their floor (30 source runs): cost against logit distance, background loss at the return (nats) and score at return (0, never-supervised level; 1, acquisition level). Tags, share of within-domain pairs ordered by cost (5,812 pairs). \textbf{e}, Residual from the single least-squares line of \textbf{b} against parameters $N$ (small, runs; large, domain means per size); line, fit over 9.7M--431M, dashed to 7B. Unit: one run.}
\end{figure}

To test whether the effect of a return is specific to its checkpoint, we added the part of its weight change that is due to the returned targets, without training, to other checkpoints of the same chess source run. At later checkpoints of 9.7M models it raised in-check legal-move mass from 0.33--0.37 to 0.61--0.62. In 64M models a change of 1.5\% of the weight norm gave 0.58--0.65 at every later checkpoint, across an 8.6-fold span of training. At earlier checkpoints it gained less, in the order of their own acquisition costs. The change made by a return of the permuted rule lost its effect on both sides of its checkpoint (Fig.~F1l--o).

\subsection{Returns at 7 billion parameters}

To test the cost relation at a larger scale, we withheld the targets that state a country's capital in 7-billion-parameter web-text models, kept every input, and returned the targets at six checkpoints in each of two seeds. From the last three checkpoints, in order, one seed reached the acquisition level with 15\%, 11\% and 1.8\% of the tokens needed from initialization and the other with 22\%, 12\% and 1.1\%. From the three earlier checkpoints neither seed reached it (Fig.~3c and Fig.~E1a,c). A permuted assignment of capitals to countries needed 12 times the tokens of the true one from the final checkpoint, and 1.3 times from initialization.

At the same logit distance, the 7B costs were close to those of much smaller models (Fig.~4a), at 0.28--1.45 times the values of the single line fitted to chess, web text and SMILES at 9.7M--431M (Fig.~4b,e). Across those sizes, the cost at a given logit distance fell by a further $0.28\pm0.09$ decades for each tenfold increase in size (Fig.~4e).

A whole domain returned in the same way. A 7B model trained with Rust documents filtered to 3\% of their share predicted Rust terms with 52\% accuracy, against 76\% for the unfiltered model at matched background loss. Continued on the unfiltered corpus, it closed half of this gap with 2.1\% of the tokens the unfiltered model had needed from initialization to reach the same accuracy (Fig.~3f and Fig.~E1j). On 212 Rust function tasks it compiled 18\% of its sampled completions at the return, against 45\% for the unfiltered model, and 37\% against 41\% after 13\% of its training tokens (Fig.~3f). Of 12 models of 20M--300M parameters trained without astronomy, Rust or Chinese documents, 11 closed half of the accuracy gap with 1.9--7.5\% of the domain tokens needed from initialization, and this share showed no trend with model size (Fig.~3d,e and Fig.~E2g).

\subsection{Removal route and earlier supervision}

We compared never-supervised source models with models from which a learned skill had been removed. In chess models from 9.7M to 163M parameters and at matched background loss, these were models whose targets were withdrawn after the skill was learned and models from which the skill was unlearned by gradient ascent~\citep{jang2023knowledge} or by misdirecting its representation~\citep{li2024wmdp}. With all targets returned, never-supervised models needed 5--20 times more tokens, depending on the route, than models that had once had the skill and started from the same legal-move probability mass (Fig.~5a,b and Fig.~F1a,c). With 10\% of the targets returned, never-supervised models reached the acquisition level in 5 of 13 cases, and models from the other routes in 51 of 52, of which 26 were at that level when the targets returned (Fig.~5c and Fig.~F1b). When training resumed without in-check targets, models unlearned by gradient ascent rose to a king-safety competence of 0.81--0.91 within 12 steps, whereas never-supervised models moved by less than 0.03 (Fig.~5d).

The size of this advantage depended on how far the removal had lowered the skill. In SMILES models with ring closure removed, never-supervised models needed 4--11 times the tokens of models that had once had the skill. In 300M web-text models, neither withdrawal nor gradient ascent brought subject--verb agreement down to the never-supervised level, and never-supervised models needed 0.8--2.7 times the tokens of the others (Fig.~F1i--k). These comparisons agree with reports of faster relearning after unlearning~\citep{ref26,lee2025distillation} and of recovery under training on unrelated data~\citep{hu2025jogging,lucki2025adversarial,siddiqui2025dormant}.

Partial supervision before the return also lowered the cost. At matched background loss, 9.7M chess models that had kept 3\% of in-check targets needed one-quarter as much training after the targets returned as models that had kept none (ratio 0.25, 95\% interval 0.21--0.31; Fig.~5e).

Related content accounted for part of the corpus support. Removing supervision on related king-safety positions, boards outside check where a move that obeys the piece-movement rules would expose the king, reduced the fitted corpus support by 63\%, whereas random removals reduced it by 33\% or 19\%, close to the shares of positions removed (Fig.~5f).

\begin{figure}[p]
\ppfigure{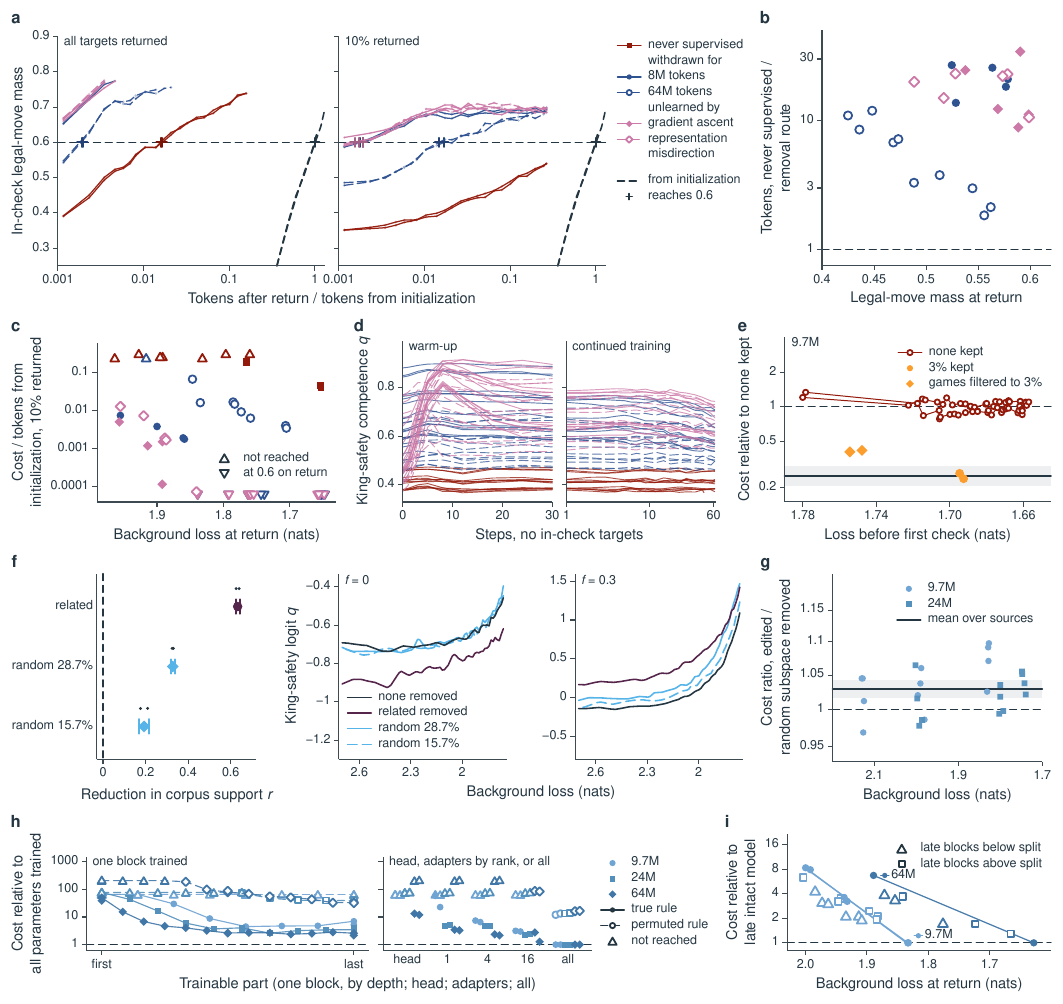}{fig5}{Fig.~5}{The removal route, earlier supervision and related content set a skill's acquisition cost.}{\textbf{a}--\textbf{d}, Chess, never-supervised models and four removal routes (key in \textbf{a}). \textbf{a}, In-check legal-move mass after all or 10\% of in-check targets return (64M, 1.89 nats, two seeds); tokens relative to training from initialization (dashed curves). \textbf{b}, All targets returned: never-supervised over removed-model tokens from the same starting mass ($n=26$). \textbf{c}, 10\% returned: cost relative to tokens from initialization, 9.7M--64M ($n=55$; triangles, not reached or at 0.6 on return). \textbf{d}, Competence $q$ during the 30 warm-up steps and continued training without in-check targets (65 models). \textbf{e}, Cost after all targets return relative to 9.7M models that kept none: kept none ($n=56$), kept 3\% ($n=2$), games filtered to 3\% ($n=2$); line and band, fitted ratio, 95\% interval. \textbf{f}, Reduction $r$ in fitted corpus support after removing related king-safety positions or random families (caps, range over seed recombinations; dots, seeds), and logit competence at $f=0$ and 0.3. \textbf{g}, Tokens to 0.6 with a rank-8 king-safety subspace removed relative to a random subspace (24 checkpoints, eight runs; line and band, mean and 90\% interval). \textbf{h}, Tokens to 0.6 with part of the network trained, relative to training all parameters from the same checkpoint: one block (by depth), the head, rank-1, -4 or -16 adapters, or all parameters; true (filled, solid) and permuted rule (hollow, dashed), 9.7M--64M (colour, marker; two source runs each); triangles, lower bounds. \textbf{i}, Tokens to 0.6 for stitched models, with blocks below a split from one snapshot and the rest from another of the same run, relative to the late intact model, against background loss at return. Filled, intact snapshots (line, least squares); open, late blocks below (triangles) or above (squares) the split (9.7M: 12 stitched, 6 intact; 64M: 6 stitched, 2 intact).}
\end{figure}

We asked how much of the network must change to acquire the skill. With a single upper block or adapters~\citep{hu2021lora} of rank 4 or 16 trainable, the return needed 1.3--7.4 times the tokens of a return that trains every parameter, which keeps 59--96\% of the saving, counted in decades (Fig.~5h). Removing the representation through which attacked squares drive king moves raised the cost by 2.9\% (Fig.~5g). Source models assembled from the lower blocks of one checkpoint and the upper blocks of another of the same run had acquisition costs between those of the two checkpoints (Fig.~5i).

\section{Discussion}

In five domains and models of up to 7 billion parameters, thinning a skill's training targets delayed the skill by an amount that small models predicted for larger ones. For king safety in chess and for capital recall at 7 billion parameters, the delay follows with no fitted parameter from a fully supervised and a never-supervised model at the same background loss; the other skills required a fitted sensitivity. Models trained without the targets had learned part of the skill from the rest of the corpus. When the targets returned, they acquired the skill with a small fraction of the data needed from initialization, and each logit of surplus lowered this cost about tenfold in chess and web text and fourfold in SMILES.

Both trends reported by Rathi and Radford~\citep{ref26} follow from these relations. Corpus support falls as a model progresses, so equation~(\ref{eq:effective-supervision}) gives a deficit that widens with progress, and larger models progress further. By the cost relation, the same progress lowers the data needed to acquire the capability once its data return. Our measurements add the size of the delay, comparisons at matched background loss and a permuted rule that tests whether the saving is specific to the withheld skill. Filtered models of 6.9 billion parameters resisted adversarial fine-tuning on 300M tokens~\citep{ref2}, 0.055\% of their pretraining tokens. In our 7B model, the filtered domain returned with up to 13\% of the model's training tokens.

For data filtering, the results separate the delay at a given background loss from the cost of acquiring the skill once its data are available. Chess models trained on games without a check ranked legal replies nearly as fully supervised models did, so a filter that removed every example of the skill left much of it to be learned from the remaining games. The cost of acquiring the skill once its data returned fell as the model progressed, and it depended on how the skill had been removed. Never-supervised chess models needed 5--20 times more tokens than models from which a learned skill had been removed. Document filtering and removing targets from the loss~\citep{ref2,ref26} therefore affect both a model's competence and the cost of adding the skill afterwards.

For evaluation, the score on a withheld skill was a weaker guide to this cost than logit distance. Longer pretraining can lower the quality a model reaches after a fixed fine-tune~\citep{springer2025overtrained}, and extended training on a sequence of tasks can reduce a network's ability to learn~\citep{dohare2024loss}. On a withheld skill, by contrast, the data needed to reach a fixed competence fell with general progress.

The change that completed a skill was small relative to the network. With a single upper block or adapters of rank 4 or 16 trainable, king safety was acquired with 1.3--7.4 times the tokens of a return that trains every parameter, and the weight change of a return raised legal-move mass at later checkpoints of the same run. This agrees with reports that pretraining lowers the number of parameters that adaptation requires~\citep{aghajanyan2021intrinsic} and that setting fewer than 0.05\% of a pretrained model's weights to zero can match full fine-tuning~\citep{lu2026deleting}.

\section{Methods}

\subsection{Domains and data}

Training corpora are contiguous token streams read once in order, so the batch schedule fixes the data order. \textbf{Chess:} rated Lichess games from 2024 between players rated at least 1,500, one move per token. \textbf{Othello:} uniformly sampled legal games~\citep{ref29} with an explicit pass token. \textbf{SMILES:} 300M canonical molecular strings from ZINC22~\citep{tingle2023zinc}, modelled at the character level. \textbf{Dyck-4:} well-formed nested sequences over four bracket pairs. \textbf{Web text:} FineWeb-Edu~\citep{penedo2024fineweb} for models of up to 163M parameters and the Dolma 1.7 recipe of DataDecide~\citep{ref42} for models of 20M--7B.

\subsection{Models and training}

All models are dense pre-norm decoder-only transformers, with width and depth scaled together across seven sizes from 0.76M to 163M non-embedding parameters, and chess models of 431M, 1.28B and 3.0B. CompleteP parameterization~\citep{ref38} ($\alpha=1$), anchored at 2.2M parameters, transfers learning rates across sizes. The main recipe uses AdamW~\citep{loshchilov2017decoupled} with a peak learning rate of 0.004, weight decay 0.1 and a warm-up--stable--decay schedule indexed in tokens per parameter. The constant-rate chess models (9.7M--3.0B) use a 15M-token warm-up and batches of 64 sequences. Sequences have 2,048 tokens. FineWeb-Edu models use a constant learning rate of 0.008 after warm-up. Dolma models have the Llama architecture as implemented in HuggingFace Transformers, without CompleteP, and use weight decay 0.05, batches of 128 sequences and, after a 32-step warm-up, a constant learning rate of $2.12\times10^{-3}$ at 20M--300M, $1.06\times10^{-3}$ at 1B and $2.65\times10^{-4}$ at 7B; the 300M and 1B models with the designs of the two 7B models (models A and B below) use the 7B rate. Seeds change the weight initialization and read the same data stream; in the Dolma models a seed also sets the data order.

\subsection{Competence, background loss and target thinning}

Target thinning removes the next-token loss at a skill's targets unless $h(p)<f$, with $h$ a deterministic hash of the stream position $p$ and $f$ the retained fraction. Inputs are never changed, and the training loss is the mean cross-entropy over the retained targets. The thinned targets are in-check boards in chess (7.1\% of tokens), forced passes in Othello (0.73\%) and ring-closing digits in SMILES (5.17\%). In web text, programmatic labellers mark closing brackets, number-marked auxiliaries, repeated 4-gram completions and counting sequences, and two factual relations, country capitals and city countries, are thinned by masking answer mentions within 128 tokens after their subject.

Competence is read on a fixed held-out split. In chess, every vocabulary token at an in-check board is assigned to an exact legality class, and king-safety competence is $q=\text{legal}/(\text{legal}+\text{king-unsafe})$ probability mass. Competence is the probability of the pass token at forced-pass positions in Othello, of the true closing digit in SMILES and, for web-text agreement, of the true form relative to the true plus the other-number form of \textit{is/are}, \textit{was/were} and \textit{has/have}. Background loss is the held-out cross-entropy, in nats, at all targets outside the thinned ones; in Othello it is taken at positions with two or more legal squares, and in the Dolma models at targets outside every labelled class. Onset is the background loss, interpolated between log-spaced checkpoints (21--36 per chess run on the warm-up--stable--decay schedule and 52--76 on the constant rate), at which competence crosses a stated level: $q=0.70$, 0.75 and 0.80 in chess, where $q=0.659$ lies on the plateau of the fully supervised trajectories, 0.4 in Othello and 0.6 in SMILES. Runs that do not reach a level are censored at their final background loss. Probes are class-balanced logistic regressions on residual-stream activations, scored against an identically probed random network, and causal use is the effect of ablating a variable's difference-of-means direction relative to random directions~\citep{ref37} (Fig.~B1i).

\subsection{Derived thinning relation}

Equations~(\ref{eq:odds}) and~(\ref{eq:effective-supervision}) are the form a weighted likelihood takes when the skill's targets enter with weight $f$ and the rest of the corpus with weight $\tau$. For each of 4,173 held-out in-check positions, the probability mass on legal and on king-unsafe moves at every snapshot of the fully supervised and the never-supervised model was interpolated, in log, to a 0.05-nat grid of background loss. The derived offset at retained fraction $f$ adds $\ln[(f+\tau)/(1+\tau)]$ to the log odds of every position of the fully supervised model and pools the masses as for $q$; $\tau$ at each background loss is the value at which $f=0$ reproduces the pooled odds of the never-supervised model. Root-mean-square differences from the measured offsets, over models and background losses, are 0.039 logits (95\% interval 0.031--0.047) for 15 thinned 9.7M models of two seeds at nine retained fractions, and 0.089 logits (0.042--0.119) for 64M models at $f=0.3$, 0.1, 0.03 and 0.01 (Fig.~1c). With each position's own never-supervised odds in equation~(\ref{eq:odds}), in place of the common $\tau$, the differences are 0.20 and 0.32 logits.

Because $q$ pools the mass of all positions, a small uniform shift of the log odds changes the pooled log odds, to first order, by the shift times $1-\operatorname{Var}(q_x)/[\bar q(1-\bar q)]$, where $q_x$ is the competence of the fully supervised model at position $x$ and $\bar q$ its mass-weighted mean; this factor is 0.59--0.66 at 1.85--2.75 nats. The pooled form, $\operatorname{logit}q(L,f)-\operatorname{logit}q(L,1)=\kappa\ln[(f+\tau)/(1+\tau)]$, is fitted with a sensitivity $\kappa$ and a corpus support $\tau$ at each background loss, to positive fractions only, over all thinned chess models (Fig.~B2k). At 64M, $\tau$ is 0.96, 0.85 and 0.80 times the 9.7M value at 1.85, 1.95 and 2.05 nats.

The same test on thinned 9.7M models of six further skills ($f=0.01$--0.5; one to three seeds per skill) gives root-mean-square errors of 0.80 logits for SMILES ring closure, 0.32 for SMILES branch closing, 0.71 for the Othello forced pass, 0.32 for the Othello forced move, 1.71 for Dyck closing brackets at match distances of 2--7 and 0.26 for web-text subject--verb agreement, against 0.31, 0.20, 0.42, 0.19, 0.39 and 0.10 for the pooled form fitted at each background loss. The derived offset is too shallow by 0.24--0.50 logits in the three skills whose never-supervised model has no competence and too deep by 0.22--0.23 in two whose never-supervised model has some; for the Dyck skill, whose never-supervised model reaches a competence of 0.27--0.43, it is too shallow by 1.36. In every skill, $\tau$ falls as background loss falls.

At 7B, three models at each of seeds 0 and 1, with one data order per seed, keep all, none and a fraction 0.3 of the capital targets: the never-supervised models follow the recipe, seed and data order of the source models of the capital returns, the fully supervised models their references, and the thinned models follow the same recipe. At each stored snapshot we scored 816 capital prompts by the probability of the correct answer among the 102 answers and interpolated the log odds of each prompt linearly in background loss to 16 losses from 3.05 to 3.80 nats, a range spanned by seven or eight snapshots of each model. The derived offset shifts the log odds of every prompt of the fully supervised model by $\ln[(f+\tau)/(1+\tau)]$, with $\tau$ the value at which $f=0$ reproduces the pooled competence of the never-supervised model of the same seed, and pools. Over the 32 offsets it differs from the measured offset by 0.14 logits (root mean square; 0.13 in seed 0 and 0.14 in seed 1), deeper by 0.10 logits on average in seed 0 and shallower by 0.09 in seed 1. The retained fraction alone ($\tau=0$) misses by 0.61 logits (0.69 and 0.52) and the offset of the never-supervised model by 0.43; with the never-supervised model of the other seed the error is 0.17 logits. Over the four lowest losses the error is 0.12 logits in seed 0 and 0.25 in seed 1, where the measured offset is deeper than the derived one and within 0.12 logits of the retained fraction alone. The measured offsets of the two seeds differ by 0.31 logits (0.44 over the four lowest losses). On country recall, whose targets all six models keep, the thinned and the fully supervised model differ by 0.09 logits. The test covers one retained fraction (Fig.~B2n).

For 18 skills of the four domains thinned to $f=0.5$, 0.3, 0.1 and 0.03 at 9.7M, the horizontal shift that aligns each thinned curve with its fully supervised curve is fitted as $A_k(1-f^p)$ (Fig.~B2c--g).

\subsection{Held-out predictions}

Each chess test fits the pooled form to calibration runs only and predicts the thinned trajectories of a held-out model from its seed-mean fully supervised trajectory, with no parameter refitted. The four tests hold out fractions below 0.1, model sizes, the constant-learning-rate schedule and the 1.28B and 3.0B models, for which the predicted fraction, $f=0.1$, is included at every calibration size. The error is the root-mean-square logit offset error of a trajectory (Fig.~B2a,b).

On web text, two sets of 24 runs vary the fractions of the four labelled classes, and of capitals and city countries, in Latin-square designs ($f=1$, 0.3, 0.1; two seeds), with five and two further runs at 1B. The response is calibrated at competence levels 0.3--0.7 with censored fits of the onset against $\ln f$ and $\log_{10}$ parameters, and only the shift in $\ln f$ enters a prediction. At 7B, model A thins closing and counting to 0.3, agreement to 0.05 and capitals to 0.1, and model B thins city countries to 0.1. Each skill is compared with the model in which it is fully supervised, whose trajectory the prediction shifts, and the error is the mean absolute difference between predicted and observed onset shift over six skills and three seeds (Fig.~1h and Fig.~B5a--c). In the isolated contrasts, only long-range closing is thinned, to 0.3, or only agreement, to 0.05, at 150M and 300M with four seeds (Fig.~B5d--i).

\subsection{Document filtering}

A chess game filter with recall $\gamma$ discards each game that contains a check with probability $\gamma$ and replaces it from the stream, which leaves a density of in-check targets of $f(\gamma)=(1-\gamma)/[0.058+0.942(1-\gamma)]$. On the constant-rate recipe, 9.7M--64M models (two seeds) were trained with game filtering or target thinning at matched residual shares (0.1, 0.03 and 0) and compared on background loss over game prefixes before the first check, with delays read at a legal-move mass of 0.6 (Fig.~1f and Fig.~B4g,h).

In web text, reference labels for astronomy (lexicon density), Rust (Rust-exclusive syntax) and Chinese (CJK share) mark 0.53\%, 0.22\% and 0.15\% of the stream. A hashed $n$-gram linear classifier over each document's first 512 tokens removes documents at thresholds that leave a share $f$ of reference-labelled tokens. Below the lowest share that a threshold alone reaches (0.1 for astronomy, 0.03 for Rust and 0.01 for Chinese), every reference-labelled document is also removed except a hashed fraction $f$ of them. Models of 20M--300M parameters were trained to 1.0--3.0B tokens in a $6\times6$ Latin square that filters all three domains in every run, with two seeds (48 runs), together with a 1B pair at $f=1$ and 0.03 and a 7B pair for Rust, model B with and without the filter at $f=0.03$. Competence is top-1 accuracy at tokens that require the domain skill in held-out domain documents (Fig.~E2).

\subsection{Chess games without a check}

The models trained on games without a check are the game-filter models at recall 1 (seed 1; 600M tokens at 9.7M--64M parameters and 450M at 163M); a replay of their training stream with python-chess finds no in-check position. They are evaluated at the first check with a played reply in each of 2,000 held-out games (1,704 positions, 1,533 with two or more legal replies), with the prompt ending at the checking move, against the model trained without in-check targets and the fully supervised model of the same size and seed at the saved checkpoint nearest in held-out loss before the first check. A board pair consists of two of these positions in which the same piece on the same square has the same two non-capturing, non-checking moves, one resolving the check on the first board and leaving the king attacked on the second, and the other the reverse (18,701 pairs from 1,162 games; intervals resample games). With each game in at most one pair, the four models preferred the legal move on both boards in 0.57--0.72 of pairs (Table~C1). For the transfer between replies, the 9.7M model was continued on 3,072 game prefixes that end at a first check, with the loss on the reply played only. Each set of replies has one run, evaluated on 326 held-out knight checks and 92 held-out positions where every legal reply is a block (Fig.~C1).

\subsection{Output classes at the skill's positions}

For 8,313 in-check positions of 2,000 held-out chess games, each of the 1,970 output tokens was assigned with python-chess to one class: legal king moves, captures of the checking piece and interpositions; moves that would be legal if the king were not in check; king moves to an attacked square; moves that expose the king to a further piece; and tokens that are not moves of the position. In the returns of Fig.~2c the model state was saved at the end of the warm-up before the return (step 0) and after steps 1--8, 10, 12 and 16 with the returned targets (2.1M tokens); over these states the share of positions whose most probable legal reply is the move played, taken as everywhere over the 7,149 positions with two or more legal replies, stayed at 0.58--0.59 for the return from 1.95 nats and at 0.62--0.64 for the return from 1.62 nats. The 23,190 SMILES ring-closure positions of Fig.~2d come from 8,000 held-out molecules.

\subsection{Acquisition cost after targets return}

\textit{Chess.} Returns start from source models trained without in-check targets (three seeds each at 9.7M, 24M, 64M and 431M) at background losses of 1.56--2.87 nats and train on unseen stream segments. The targets return at a share $s=1$, 0.3, 0.1 or 0.03 with 131,072 tokens per step, in one of two return conditions. With the warm-up before the return, the AdamW second moments are estimated over 20 steps without parameter updates, the learning rate is warmed up over 30 steps without in-check targets, and the targets then return at full learning rate (110 runs). Costs count the tokens after the targets return; with these 50 steps also counted, the costs from each source run's lowest-loss checkpoint are 3.1--7.5\%. With the warm-up at the return, a 20-step warm-up starts when the targets return (188 runs; Fig.~D1a--d). Six returns of all targets from 9.7M and 64M checkpoints at 2.99--3.99 nats use the warm-up before the return. Competence and background loss are evaluated on 1,000 held-out games when the targets return and then at steps spaced by a factor of 1.25. The cost is the number of tokens from the return to the first evaluation at or above the acquisition level, interpolated from the evaluation before it, divided by the tokens a same-size model needs from initialization. Acquisition is read on the in-check legal-move probability mass at a level of 0.6, and king-safety competence $q$ is used for the thinning results; at the first crossing of $q=0.659$ in a return, the legal-move mass has a median of 0.60.

\textit{Web text.} Eleven source models of 9.7M--163M parameters were trained on FineWeb-Edu for 4--64 tokens per parameter (64, 24 and 12 at 9.7M, 24M and 64M; 8 and 4 at 163M) without the targets of number-marked finite verbs. At a return the AdamW moments are estimated as in chess, 30 steps on the source model's own targets follow, and all agreement targets then return at 64 sequences per step on unseen rows (90 runs). Competence is evaluated at auxiliaries with a noun subject in 384 held-out sequences and the acquisition level is 0.85. The denominator is the mean over runs from initialization at the same size, recipe and batch.

\textit{SMILES.} Ring-closing targets return to the 9.7M, 24M, 64M and 163M source models, one seed each, at seven to nine checkpoints per size with background losses of 0.48--0.85 nats (21, 19, 18 and 15 runs), at a share of 1, 0.3, 0.1 or 0.03 (64M: 1, 0.3 or 0.1; 163M: 1 or 0.3) and 16 sequences per step, after 20 moment-estimation steps and with the warm-up at the return. Competence is evaluated on 8,000 held-out molecules, the acquisition level is 0.6, and budgets are 20M--100M tokens by share. The denominator is the run from initialization of the same size and seed with the same batch and warm-up (11.3M, 11.6M, 15.8M and 20.7M tokens).

\textit{Othello.} All forced-pass targets return to eight source runs trained with none, 10\% or 30\% of them (53 runs), with the warm-up at the return and 64 sequences per step. Competence is evaluated on 2,000 held-out games, the acquisition level is 0.4 and the budget is 40M tokens. The denominator is the mean over the fully supervised runs of the same size.

\subsection{Logit distance and the cost relation}

For each return, the acquisition loss $L^*(s)$ is the background loss at which the reference for the returned share $s$ reaches the acquisition level. For the progress term and the split of the cost relation at $L^*(s)$, it is read from one curve per domain, fitted to the references across sizes and returned shares in chess, across returned shares at 9.7M in SMILES and across sizes in web text, which lies within 0.04 nats of the same-size reference in chess and SMILES and within 0.08 nats in web text. The reference is the model of the same size and recipe trained from initialization with a fraction $s$ of the skill's targets, of the same seed as the source run where its trajectory extends to $L_0$ and of another seed otherwise. The logit distance is $x=\ell^*-\ell_s(L_0)$, with $\ell^*$ the log odds of the acquisition level and $\ell_s(L)$ the reference log odds, a centred running mean over three checkpoints (five at 7B). $L_0$ is the background loss at the first evaluation of the return run, when the targets return: after the 30 preceding steps in chess with the warm-up before the return and in web text, and at the stored checkpoint with the warm-up at the return (chess, SMILES and Othello).

Acquisition cost is modelled as $P+10^{\,a+b\min(x,0)}$. The progress term $P$ is the number of tokens the source run needs to bring its own background loss from $L_0$ to $L^*(s)$, divided by the tokens from initialization; it is zero when $L_0\le L^*(s)$ and has no parameter fitted to the costs. The main-text rates are from a maximum-likelihood fit with an intercept $a$, a rate $b$ and a residual variance for each domain and a random intercept for each source run, over the 312 runs with a reference that reached the acquisition level in chess (159 runs, 12 source runs), web text (88, 11) and SMILES (65, 4) at 9.7M--431M parameters. The rates differ between domains (likelihood ratio $\chi^2=44.4$, 2 degrees of freedom). The single line of Fig.~4b is the least-squares fit of one intercept and one rate to the same runs, $a=-0.53$ and $b=0.97\pm0.07$, with a residual standard deviation of 0.36 dex. The least-squares slope of its residuals against $\log_{10}$ parameters is the trend with model size (Fig.~4e). The 7B costs are not in the fit: the measured costs are 0.45--1.02 times the line's values in seed 0 and 0.28--1.45 times in seed 1, and the six censored returns have lower bounds 0.21--0.43 dex below $P+10^{a}$. Othello is not fitted: in 46 of its 53 runs the level was reached within five optimizer steps, the resolution of the cost measure.

A source model is near the floor when its score lies in the lowest 10\% of the range between the median score of never-supervised models, taken above the acquisition loss, and the acquisition level; in chess this score is $q$, with the level of 0.659. Near-floor ordering is the share of pairs of runs within a domain whose order of cost is identified and agrees with the order of the predictor: 0.85 for logit distance, 0.77 for background loss and 0.53 for the score when the targets return (Fig.~4d).

\subsection{Batch size and learning rate at the return}

Chess source models at 9.7M, 24M and 64M parameters received all in-check targets, with the warm-up before the return, from three checkpoints each, with batches of 16, 64 or 256 sequences at the standard learning rate and of 64 or 16 sequences at one third of it. The warm-up used the setting of the return, every setting read the same token stream in two data orders, and each cost is divided by a run from initialization at the same setting. Because the warm-up moves the background loss by an amount that depends on the setting, the logit distance is read here at the stored checkpoint. The slope of $\log_{10}$ relative cost against logit distance, pooled over sizes, was $1.03\pm0.07$ decades per logit at the standard setting (18 runs) and 0.86--1.29 at the other four. From the lowest-loss checkpoint at 9.7M, the permuted rule needed 11.0--12.7 times the tokens of the true rule at the same settings (Fig.~D1i--k).

\subsection{Transplanted weight changes}

In the two 9.7M chess source runs, the targets were returned at the 619M, 290M, 193M and 97M-token checkpoints with the warm-up before the return, each return was stopped at the first evaluation at which in-check legal-move mass reached 0.6, and its weight change was saved. A control run from each checkpoint took the same number of steps on the source's own targets, and the change due to the returned targets is the difference between the two weight changes. It was added, without training, to each of the 22 saved checkpoints of the same run and of the other seed's run. The 193M change gave a legal-move mass of 0.59 at its own checkpoint, 0.61--0.62 at every later one with a rise in background loss of 0.04--0.05 nats, and 0.54 and 0.45 at 145M and 97M tokens. Across the checkpoints with a measured acquisition cost, the rank correlation between the recipient's cost and the mass gained from the 290M change was $-0.95$ and $-0.77$ in the two seeds. The changes gave no gain at the other seed's checkpoints, and neither did a Gaussian change with the norm of the return's whole weight change in every tensor. The weight change of a permuted-rule return at 619M tokens, read out on the permuted rule, gave 0.61 at its own checkpoint and 0.41, 0.30, 0.16 and 0.11 at the four earlier ones. In two 64M source runs with 31 checkpoints each, the 226M change, 1.5\% of the weight norm, gave 0.58 at the next checkpoint and 0.63--0.65 from 322M to 1.93B tokens with a rise in background loss of 0.01--0.02 nats, and the permuted-rule change from 226M tokens gave 0.62 at its own checkpoint, 0.37 at the next and 0.07 at the last (Fig.~F1l--o).

\subsection{Returned targets at 7B}

The 7B source model follows the recipe of models A and B with the targets of country capitals removed from the loss, together with agreement and repeated 4-gram targets, and every input kept, in seeds 0 and 1. Six snapshots between steps 2,048 and 12,817 are continued on the source model's own data order. The capital targets return with the warm-up before the return, during which the source model's masks apply. The acquisition level is a capital competence of 0.20; the fully supervised references end at 0.33 and 0.39. The reference of each seed is a fully supervised model with the recipe, data order and seed of its source model and the capital targets kept; it reaches the level after 1.67B tokens in seed 0 and 1.63B in seed 1, at a background loss of 3.22 nats in both. Capital competence is smoothed over five evaluations in logit, and the level is reached at the first upward crossing of the smoothed curve. The returns from steps 12,817, 8,192 and 5,793 reached the level after 1.8\%, 11.0\% and 15.2\% of the reference's tokens in seed 0 and after 1.1\%, 12.3\% and 22.3\% in seed 1, and those from the three earlier steps did not reach it in either seed. The returns from step 12,817 start 0.013 and 0.017 nats below the lowest background loss of their reference, so their logit distance is read on model B of seed 0. The permuted assignment maps the first answer token of each capital through one fixed derangement of the 102 answer groups, and a matched true-rule control supervises the same positions with the true first token; both were run in seed 0 only. Agreement targets returned to the same six snapshots reached a competence of 0.85 after 2.7--5.6\% of the tokens needed by model B of the same seed, which keeps the agreement targets (Fig.~E1c,e). With only this token supervised, the permuted assignment needed 12.3, 3.9 and more than 1.4 times the tokens of the true assignment from steps 12,817, 8,192 and 4,096. A 7B model of seed 0 that keeps every target of the recipe, with the capital targets permuted from initialization, reached the level after 1.29 times the tokens of the seed-0 reference.

\subsection{Domains returned after filtering}

The Rust-filtered 7B model after 15,000 steps is continued with its optimizer state on the unfiltered corpus, at the same learning rate and batch, for 0.5B tokens on data blocks that neither it nor model B has read, and model B, the unfiltered model of Fig.~3f, is continued from its own state on the same blocks. Term accuracy on the Rust evaluation documents is smoothed over five evaluations in logit, and the gap is the mean of model B's first five evaluations minus the filtered model's accuracy at the return (0.764 and 0.517). For each fraction of the gap, the tokens after the return are divided by the tokens at which model B, evaluated on the same documents from initialization, first reached that accuracy. Half of the gap closed after 9.4M tokens against 450M from initialization, three quarters after 3.4\% and nine tenths after 13\%. Before the return the filtered model had read 259,000 reference-labelled Rust tokens, the Rust content of 117M tokens of the unfiltered stream; with these counted, its Rust tokens at the three points are 28\%, 13\% and 18\% of those model B had read at the same accuracy. The data order of the 7B models reads rows that hold 55 of the 290 Rust evaluation documents. On the 235 that it does not read, the single-evaluation accuracies of model B and the filtered model at the return are 0.746 and 0.509, against 0.753 and 0.517 on all documents, and the fraction of the gap closed at each of seven saved states of the continuation is within 0.02 of its value on all documents. Working behaviour is scored on 212 Rust function tasks: ten completions per task are sampled at temperature 0.8, compiled with the hidden assertions and run in a sandbox, and compile and pass rates are fractions of the 2,120 completions. At the return the filtered model compiled 17.6\% (95\% interval 15.0--20.2) and model B 44.7\% (40.4--49.1); after 500M tokens the values were 37.4\% (33.2--41.4) and 41.1\% (37.0--45.3).

At 20M--300M, models trained with one domain removed ($f=0$, seed 0) were continued on the unfiltered stream for 300 steps of 128 sequences, with a new data order and a 32-step learning-rate warm-up, and evaluated every 10 steps (Fig.~E2g). Shares follow the 7B rule, in reference-labelled domain tokens. The $\log_{10}$ shares were fitted by maximum likelihood with censoring; the Chinese model at 300M is a lower bound (5.9\%). A constant, $-1.33\pm0.05$ (4.7\%) with a residual standard deviation of 0.17 dex, was preferred to a form with a size term ($+0.02\pm0.11$ dex per tenfold increase) and gives a 95\% prediction interval of 1.9--11.9\% at 7B (Fig.~3e).

\subsection{Removal routes}

King safety is removed by four routes in 13 cases, each pairing a fully supervised chess model with a model of the same seed and training tokens trained without in-check targets, at 9.7M, 24M, 64M and 163M parameters and background losses of 1.65--1.96 nats, where the models of a case agree within 0.015 nats. The two withdrawn routes continue the fully supervised model without in-check targets for 8M and for 64M tokens; when the targets return, the models withdrawn for 64M tokens are up to 0.12 nats below the never-supervised model of their case and the others within 0.04. Unlearning either trains by gradient ascent on the in-check loss (weight 0.3) with descent on the remaining loss~\citep{jang2023knowledge}, or pushes the residual stream at in-check positions towards a fixed random direction while holding it to its value before unlearning elsewhere~\citep{li2024wmdp}; both keep the state of lowest competence whose background loss is at most 0.05 nats above the start. Every model then starts from a fresh optimizer and receives the in-check targets with the warm-up before the return, at a share of 1, 0.1 or 0. The matched comparison enters the never-supervised trajectory of the same case at the legal-move probability mass from which the other model starts and compares the tokens each then needs to reach a mass of 0.6.

On web text the routes are applied to subject--verb agreement in 300M models of the Dolma recipe, two seeds at two background losses each. Gradient ascent uses the two-way cross-entropy of the auxiliary against its number partner (weight 0.3, learning rate $10^{-4}$, 64 sequences per step); with ascent on the full cross-entropy, as in chess, competence stayed above 0.87 (Fig.~F1i--k). Withdrawal lowered competence to 0.65--0.74 and two-way ascent to 0.61--0.70, against 0.50--0.51 for never-supervised models, so the matched comparison of chess is made on competence ($n=7$). Returns follow the chess procedure with 64 sequences per step, the acquisition level is 0.85, and costs are divided by the tokens at which the fully supervised run of the same seed reaches it.

In SMILES the routes are applied to ring closure in 9.7M and 24M models, two seeds each, at 6 and 14 tokens per parameter (eight cases), with the chess settings and 16 sequences per step. Withdrawal lowered competence from 0.86--0.91 to 0.01--0.04, gradient ascent stopped at 0.16--0.47 and misdirection at 0.09--0.14, against at most 0.002 for never-supervised models.

\subsection{Earlier supervision}

Two 9.7M chess models that kept 3\% of in-check targets, and two trained on games filtered to a residual share of 3\%, received all in-check targets with the warm-up before the return. The reference is the set of such returns to models trained without in-check targets, whose $\log_{10}$ cost is fitted as a cubic in background loss before the first check with an intercept for each size. The effect of earlier supervision is the mean, over the two models that kept 3\%, of each model's mean residual from the reference fit, and the ratio of 0.25 is ten to the power of this effect. Its 95\% interval uses the variance between and within source runs of the reference, estimated with each source run left out in turn (Fig.~5e).

\subsection{Related king-safety positions}

Family-masking runs in 9.7M chess models mask the targets on related king-safety boards, or on random families of 28.7\% and 15.7\% of positions, at in-check retained fractions $f=0$, 0.03, 0.3 and 1 with two paired seeds. For each seed, the models without a masked family at the four fractions, $f=0$ included, give $\kappa(L)$ and $\tau(L)$ at nine background losses from 1.90 to 2.10 nats. With these fixed, the offsets of the masked from the unmasked models are fitted by least squares with $D_k(f,L)=c_k+\kappa(L)\ln\{[f+\tau(L)(1-r_k)]/[f+\tau(L)]\}$, where $c_k$ does not depend on $f$ and $r_k$ is the fractional reduction in corpus support. The reported reduction is the mean of the two seed fits, and the range is over the 16 assignments of a seed to each fraction (Fig.~5f).

\subsection{Restricted returns, representation edit and stitched models}

Restricted returns start from the lowest-loss checkpoint of two chess source runs at each of 9.7M, 24M and 64M parameters and use the warm-up before the return with only selected parameters trainable: one transformer block, the final norm and unembedding, or rank-1, -4 or -16 low-rank adapters~\citep{hu2021lora} on every block. Of three learning-rate multipliers (one, three and ten), the one with the cheapest permuted return was used for both rules. Costs are tokens to an in-check legal-move mass of 0.6 divided by those of the unrestricted return from the same checkpoint, and a run that does not reach 0.6 within half the tokens from initialization is a lower bound. The share of the saving kept is $1-\log_{10}(\text{cost ratio})/\log_{10}(\text{tokens from initialization}/\text{tokens of the unrestricted return})$. Where both rules reached 0.6 under a restriction, the permuted rule needed 10--66 times more tokens.

For the representation edit, distributed alignment search fitted a rank-8 subspace at the token where the move is predicted, in the block with the highest held-out interchange transfer, at three checkpoints of each of eight chess source runs (9.7M and 24M parameters, four seeds each). The readout is king-move sensitivity, $\log p(K\to s\mid s\ \text{safe})-\log p(K\to s\mid s\ \text{attacked})$, on counterfactual position pairs from held-out games. The edit replaces the subspace by its mean at every token and removed 50--64\% of the readout. Each checkpoint was continued with the edit or with a random rank-8 ablation, kept in place while the targets returned. The contrast is the paired difference in $\log_{10}$ tokens to a legal-move mass of 0.6, averaged over a source run's three checkpoints, with a 90\% $t$ interval over the eight source runs (Fig.~5g and Fig.~F1f--h).

Stitched models take the embedding and the blocks below a split $K$ from one checkpoint and the remaining blocks, final norm and unembedding from another checkpoint of the same chess source run, and receive the targets with the warm-up before the return in two data orders. Their costs are compared with the least-squares line of log cost against background loss at the return through the intact models of the same run (Fig.~5i).

\subsection{Statistics}

The unit of replication is a training run from one initialization seed. The thinned models of the controlled domains have three to five seeds per size up to 64M parameters and one to three at 163M and 431M; designs with one or two source runs per size are stated in their sections. Web text has two seeds per design at 300M and 1B, three at 7B and four per size in the isolated contrasts; the capital returns at 7B have two seeds, and the document-filter pair at 1B, the Rust-filtered 7B model and the domain returns at 20M--300M are single runs at seed 0. The chess models trained on games without a check have one seed per size, and each set of replies in the transfer test has one run. Checkpoints of a run, and the returns that start from them, are repeated measurements of that run, so cost fits include a random intercept for each source run or standard errors clustered by source run. A value after $\pm$ is a standard error, and intervals are 95\% unless stated. Errors of the derived relation have intervals from resampling models. A run that does not reach a level within its budget is a lower bound at the tokens it ran, and a run at the level by its first evaluation is an upper bound. Bounds enter an ordering only where the order of the pair is identified, enter fits by maximum likelihood with censoring where stated, and are left out of least-squares fits of cost, as are runs without a reference trajectory and the Othello runs.

\subsection{Software}

Experiments and analyses used PyTorch 2.12, NumPy 2.5, SciPy 1.18, python-chess 1.11.2, RDKit (2021.03 for corpus generation and 2026.09 for evaluation), SELFIES 2.2.0, Numba 0.68, spaCy 3.8.7 and HuggingFace tokenizers 0.23 and Transformers 5.19. Figures were rendered in inklet 4.9; chess pieces use the vector set by Colin M. L. Burnett distributed with python-chess. Large language model assistants (Anthropic Claude, Google Gemini and OpenAI Codex) were used under the authors' direction to assist with training and analysis code, checks of the labellers and manuscript editing; the authors verified every reported result and take full responsibility for the content.

\subsection{Data availability}

All training corpora are public or generated from a seed. Chess games are the rated games of January--August 2024 in the Lichess database (\url{https://database.lichess.org}). SMILES strings are the first 300M molecules of the Hugging Face dataset \texttt{chandar-lab/ZINC\_22}, without 830 that fail a SELFIES round trip, and the held-out molecules are from the last 3M of the same split. Web text is from \texttt{HuggingFaceFW/fineweb-edu} and from the tokenized Dolma 1.7 files of \texttt{allenai/DataDecide-data-recipes}. Othello games and Dyck-4 sequences are generated from a seed. The released code rebuilds the training streams, label arrays and held-out sets from these sources; its SMILES stream keeps the 830 molecules, so stream positions after the first of them differ from ours.

\subsection{Code availability}

Code for building the corpora, training with target thinning, returning targets, the removal routes, document filtering and the fits of the thinning and cost relations is available at \url{https://github.com/Mapika/progress-substitutes-data} under the MIT licence. Its README lists the analyses that the release leaves out, among them the Rust function tasks and the transplanted weight changes, and the points at which a rerun differs from our runs.

\clearpage
\appendix
\renewcommand\theequation{\thesection\arabic{equation}}\renewcommand\theHequation{app\theequation}
\phantomsection\pdfbookmark[0]{Appendices}{appendices}
\ppappendicestitle{Appendices}\markboth{Appendices}{}
\noindent Appendix~A derives the thinning relation. Appendices~B--G follow the order of Methods and give the detail that Methods leaves out, each with its tables and figures.

\begingroup\pdfbookmark[1]{Contents}{appcontents}
\ppcontentsgroup{Contents}
\ppcontentsentry{A}{Derivation of the thinning relation}{app:A}
\ppcontentsentry{B}{Target thinning and the delay}{app:B}
\ppcontentsentry{C}{Outputs of a never-supervised model}{app:C}
\ppcontentsentry{D}{Acquisition cost and the cost relation}{app:D}
\ppcontentsentry{E}{Returns at 7 billion parameters and of whole domains}{app:E}
\ppcontentsentry{F}{Removal routes, earlier supervision and restricted returns}{app:F}
\ppcontentsentry{G}{Statistics}{app:G}
\endgroup

\section{Derivation of the thinning relation}\label{app:A}\setcounter{equation}{0}

\textit{Setting.} A training position $x$ has a played move $y$. The skill's targets are the targets at positions of one class, in chess the positions with the side to move in check. Under target thinning each of these targets enters the loss with probability $f$ and every other target with weight 1, so a skill target has expected weight $f$. At a held-out position $x$ of the class, $\ell_x$ is the probability mass on legal moves, $u_x$ the mass on king-unsafe moves, $O_x=\ell_x/u_x$ the odds of a legal escape and $q_x=\ell_x/(\ell_x+u_x)$ the competence at that position.

\textit{Weighted likelihood.} The derivation rests on three assumptions. First, at background loss $L$ the prediction at $x$ depends on $x$ through a representation that does not depend on $f$. Second, for each value of the representation the prediction is the weighted-likelihood optimum, the weighted frequency of the training targets that share it. Third, the skill's targets that share the representation of $x$ place no mass on the king-unsafe moves of $x$. Among the targets that share the representation of $x$, let $\pi_S$ be the distribution of the skill's own targets, $\pi_R$ the distribution of the targets from the rest of the corpus, and $\tau_x$ the weight of the rest of the corpus per unit weight of the skill's own targets. The second assumption gives the prediction for a move $m$,
\begin{equation}
p_f(m\mid x)=\frac{f\,\pi_S(m)+\tau_x\,\pi_R(m)}{f+\tau_x}.
\label{eq:s-mixture}
\end{equation}

\textit{Odds.} Write $\ell_S$ for the mass of $\pi_S$ on the legal moves of $x$, and $\ell_R$ and $u_R$ for the masses of $\pi_R$ on its legal and king-unsafe moves. With the third assumption, equation~(\ref{eq:s-mixture}) gives
\begin{equation}
O_x(f)=\frac{f\,\ell_S+\tau_x\,\ell_R}{\tau_x\,u_R}=O_x(0)+f\,\bigl[O_x(1)-O_x(0)\bigr],
\label{eq:s-odds}
\end{equation}
which is equation~(\ref{eq:odds}) of the main text: the odds are linear in the retained fraction between the never-supervised model ($f=0$) and the fully supervised model ($f=1$).

\textit{Shift at one position.} Dividing equation~(\ref{eq:s-odds}) by $O_x(1)$ gives the change in log odds,
\begin{equation}
\operatorname{logit}q_x(f)-\operatorname{logit}q_x(1)=\ln\!\left(\frac{f+\tau'_x}{1+\tau'_x}\right),\qquad \tau'_x=\tau_x\,\frac{\ell_R}{\ell_S},
\label{eq:s-shift}
\end{equation}
which is equation~(\ref{eq:effective-supervision}) with corpus support $\tau'_x$. At one position the log odds follow the logarithm of effective supervision with a sensitivity of one.

\textit{Corpus support from the never-supervised model.} At $f=0$, equations~(\ref{eq:s-odds}) and~(\ref{eq:s-shift}) give
\begin{equation}
\frac{\tau'_x}{1+\tau'_x}=\frac{O_x(0)}{O_x(1)}.
\label{eq:s-tau}
\end{equation}
Corpus support is therefore the odds of the never-supervised model in units of what the skill's own targets add. It is read from a model trained with none of the targets and a fully supervised model at the same background loss, and no thinned model enters. Only the expected weight of a target enters equation~(\ref{eq:s-mixture}), so a loss weight of $f$ on every target and a fraction $f$ of the targets kept at full weight give the same prediction; the measured difference between the two at $f=0.1$ is 0.009 logits (Fig.~B2h).

\textit{Pooling over positions.} Competence $q$ pools the masses of all test positions, $q=\sum_x\ell_x/\sum_x(\ell_x+u_x)$. With $m_x=\ell_x+u_x$, this is the mass-weighted mean $\bar q=\sum_x m_x q_x/\sum_x m_x$. When the log odds of every position shift by the same amount $\delta$ at fixed $m_x$, each $q_x$ changes at the rate $q_x(1-q_x)$, and the pooled log odds change at the rate
\begin{equation}
\frac{\mathrm{d}\operatorname{logit}\bar q}{\mathrm{d}\delta}=\frac{\operatorname{E}_m[q_x(1-q_x)]}{\bar q\,(1-\bar q)}=1-\frac{\operatorname{Var}_m(q_x)}{\bar q\,(1-\bar q)},
\label{eq:s-pool}
\end{equation}
where the mean and the variance are taken over positions with weights $m_x$. The pooled sensitivity is below one by the spread of competence across positions, a quantity of the fully supervised model alone. The pooled form used for fits is
\begin{equation}
\operatorname{logit}q(L,f)-\operatorname{logit}q(L,1)=\kappa(L)\,\ln\!\left(\frac{f+\tau(L)}{1+\tau(L)}\right),
\label{eq:s-pooled-form}
\end{equation}
with a sensitivity $\kappa$ and a corpus support $\tau$ at each background loss.

\textit{Derived offsets.} The derived offset at retained fraction $f$ applies equation~(\ref{eq:s-shift}) with one $\tau(L)$ to every position of the fully supervised model and pools the masses as for $q$; $\tau(L)$ is the value at which $f=0$ reproduces the pooled odds of the never-supervised model. The relation holds for retained fractions between 0 and 1.

\textit{Measurements.} In chess, equation~(\ref{eq:s-pooled-form}) fitted to the median over positions of the log-odds difference from the fully supervised model gives a sensitivity of 0.95--1.17 and 1.02--1.20 for the two 9.7M seeds at 20 background losses, and of 0.99--1.09 at 64M, as equation~(\ref{eq:s-shift}) states. Equation~(\ref{eq:s-pool}), evaluated on the fully supervised 9.7M model, is 0.59--0.66 between 1.85 and 2.75 nats, where the sensitivity fitted to the pooled offsets is 0.55--0.65; at 64M it is 0.67, 0.63 and 0.62 at 1.85, 1.95 and 2.05 nats. Corpus support measured from the two 9.7M seeds is 0.143 and 0.157 at 2.75 nats, 0.077 and 0.075 at 2.35 nats, 0.028 and 0.026 at 1.95 nats and 0.015 and 0.014 at 1.85 nats. It falls because the odds of the fully supervised model grow fivefold between 2.8 and 1.85 nats while those of the never-supervised model grow 1.5-fold. At 64M it is 0.014, 0.023 and 0.031 at 1.85, 1.95 and 2.05 nats, against 0.014, 0.027 and 0.039 at 9.7M.

The derived offsets differ from the measured ones by 0.039 logits (root mean square; 95\% interval 0.031--0.047) over 300 points from 15 thinned 9.7M models, and by 0.089 logits (0.042--0.119) over 20 points from four 64M models. Equation~(\ref{eq:s-pooled-form}) with both parameters fitted at each background loss has errors of 0.028 and 0.094 logits on the same models, and an offset of $\ln f$, the relation without corpus support, has errors of 0.92 and 1.59 logits. At 9.7M the mean measured and derived offsets are $-1.28$ and $-1.22$ logits at $f=0.05$, $-1.07$ and $-1.03$ at 0.1, $-0.81$ and $-0.78$ at 0.2, $-0.62$ and $-0.61$ at 0.3, $-0.36$ and $-0.37$ at 0.5 and $-0.20$ and $-0.20$ at 0.7. At 64M they are $-0.73$ and $-0.71$ at $f=0.3$, $-1.32$ and $-1.27$ at 0.1, $-1.88$ and $-1.75$ at 0.03 and $-2.12$ and $-2.02$ at 0.01.

\clearpage
\section{Target thinning and the delay}\label{app:B}\setcounter{equation}{0}

\subsection{Domains and data}
Each controlled domain provides exact labels for its latent variables at every prefix, together with a deterministic verifier. Othello and Dyck-4 have an exact Bayes-optimal predictor, which gives the floor of the cross-entropy.

\subsection{Models and training}
The seven base sizes have 0.755M, 2.21M, 4.82M, 9.67M, 24.4M, 64.4M and 162.8M non-embedding parameters. FineWeb-Edu models train with weight decay 0.1 and gradient clipping at 1. On the Dolma recipe the learning rate at 1B is half that of the 20M--300M models, and the learning rate at 7B was selected by the lowest loss after 700 steps among three rates. Models A and B stop at 3.4B and 3.9B tokens.

\subsection{Competence, background loss and target thinning}
Masked positions are excluded from the numerator and the denominator of the training loss, so every other target keeps its loss term and its weight relative to the thinned targets changes. A run that does not reach an onset level is censored at its final background loss, which is an upper bound on the background loss at which the level would be crossed.

\textit{Background loss as the index (Fig.~B1).} For each skill, the logit competence of fully supervised runs is fitted as a cubic in background loss (fifth degree for web text) on seed 0 of the smaller sizes, and the other seeds and the larger sizes are scored from that fit. Over the 98 runs left out of the fits, across 21 skills, the median absolute residual is 0.031 logits, and it is at most 0.16 logits for the 21 runs of 100M parameters or more.

\textit{Loss weight against thinning (Fig.~B2h).} In 9.7M chess models, two seeds, every in-check target is given a loss weight of 0.1, or 10\% of in-check targets are kept at full weight. Both divide the summed loss by the number of eligible targets. The weighted minus the thinned logit competence at matched background loss, between 1.85 and 2.80 nats, is $+0.009$ logits ($+0.002$ and $+0.016$ in the two seeds).

\textit{Delay curve and size term (Fig.~B3).} The pooled delay curve is a least-squares fit of $\Delta L=b\,(1-f^{c})$ to the 50 thinned chess runs with an onset at $q=0.75$, across all sizes and both schedules, with $b=0.77\pm0.07$, $c=0.28\pm0.03$ and a root-mean-square error of 0.020 nats. For the warm-up--stable--decay runs, tokens to onset are read by linear interpolation between snapshots at the background loss of onset. The size term is the slope of the delay against $\log_{10}$ parameters at a fixed fraction, pooled over fractions. On the constant learning rate it is $0.033\pm0.006$, $0.014\pm0.007$ and $0.002\pm0.007$ nats per tenfold increase in size at $q=0.70$, 0.75 and 0.80, and on the warm-up--stable--decay schedule $-0.003\pm0.009$ and $-0.010\pm0.011$ at $q=0.75$ and 0.80. At $q=0.659$, on the plateau of the fully supervised trajectories, it is $0.062\pm0.009$ and $0.040\pm0.010$ on the two schedules. Onsets are therefore read at $q\ge0.70$.

\subsection{Derived thinning relation}
Appendix~A gives the derivation. Errors of the derived offsets are taken over 1.85--2.80 nats at 9.7M and 1.85--2.05 nats at 64M. The sensitivity fitted to the pooled offsets at each background loss, across 2.2M--431M and both schedules, is 0.55--0.67, with $\kappa$ and $\tau$ fitted within a shared window of background loss and $\kappa$ resampled across runs. The slope of pooled logit competence against $\ln f$ is $\kappa f/(f+\tau)$, so $\tau$ is the fraction at which the response to proportional thinning halves. Fitted across sizes with the pooled sensitivity, $\ln\tau$ changes by $+0.03$ for each tenfold increase in parameters (95\% interval $-0.05$ to $+0.09$) over 2.2M--431M at fixed background loss. In the fit of the horizontal shift, $A_k(1-f^p)$, each of the 18 skills has its own amplitude $A_k$ and the exponent $p$ is shared; the fit is tested on another seed and on eight skills at 24M (Fig.~B2c--g).

\subsection{Held-out predictions}
The four chess tests hold out fractions below 0.1 (28 trajectories in 8 groups that share a fully supervised trajectory), model sizes (99 in 12), the constant-learning-rate schedule (17 in 5) and the 1.28B and 3.0B models (2 in 2), and the error is pooled within each test.

On web text, onsets are read against the held-out background loss outside all thinned targets, and the response is calibrated with censored fits of
\begin{equation}
\text{onset}=g(f)+d\,\log_{10}\!\left(\frac{N}{\bar N}\right)+e_{\text{cell}}+e_{\text{seed}},\qquad g(f)=g_1+a\ln f,
\label{eq:s-separable}
\end{equation}
where $N$ is the number of parameters, $\bar N$ the geometric-mean size, $d$ the size term and $e_{\text{cell}}$ and $e_{\text{seed}}$ random effects. Nested closing is an evaluation subset of the closing mask. Neither 7B model is fully supervised on every skill, so each skill is compared across the two models, which differ in several thinned families at once; the isolated contrasts change one family at a time. The error of 0.058 nats is the mean, over the six skills and the three seeds, of each seed's mean absolute difference between predicted and observed onset shift at competence levels 0.3--0.7 (0.055, 0.061 and 0.057 for the three seeds). The same two designs give 0.067 nats at 300M and 0.063 at 1B, with two seeds each.

In the isolated contrasts every other family is fully supervised, and within a seed the three runs share initialization, data order and mask hashes. Runs train to 2B tokens at 150M and 3B at 300M, and the agreement-thinned runs to 1B. The 150M offset, as a function of background loss, is added to each 300M seed's fully supervised trajectory with a 95\% predictive band of $\pm t_{0.975,3}\,\hat\sigma\sqrt{1.25}$ from the four 150M seeds, and beyond the 150M range of background loss the offset at the edge is held.

\subsection{Document filtering}
In chess, 163M models were trained at $f=1$ and 0. The delay ratio at a residual share is the delay of the filtered model divided by the delay of the model with in-check targets thinned to the same share, both read from the interval regression of onset on the loss before the first check at a legal-move mass of 0.6. It is 0.87 at a share of 0.1 (0.84--0.90 over sizes and seeds) and 0.78 at 0.03 (0.77--0.80). In SMILES, removing whole molecules brings ring closure 0.19 nats earlier than masking its targets when both are read on the shared background loss, and 0.05 nats later on each run's own training mixture (Fig.~B4e).

In web text, the Chinese label is the share of CJK characters, excluding kana and hangul. The classifiers have average precisions of 0.88, 0.98 and 0.98 for astronomy, Rust and Chinese, and at $f=0$ every reference-labelled document is removed as well. The Latin-square runs are accompanied by runs with thinned domain targets and with a random residual (Fig.~E2b). The realised share of the 7B Rust filter is 0.029. The background loss of the document-filtering runs leaves out the held-out targets in documents that resemble the three domains, about 2.6\% of the targets. Onsets are read at half of each domain's competence range. The effect of each filter on the other two domains is the onset fitted for a domain at its own share and size minus the observed onset, plotted against the share kept by the other filter; the means lie within 0.03 nats of zero (Fig.~E2e). The pair of 1B models filtered to $f=0.03$ delays astronomy, Rust and Chinese by 0.62, 0.97 and 0.29 nats.

\clearpage
\ppappfigure[44pt]{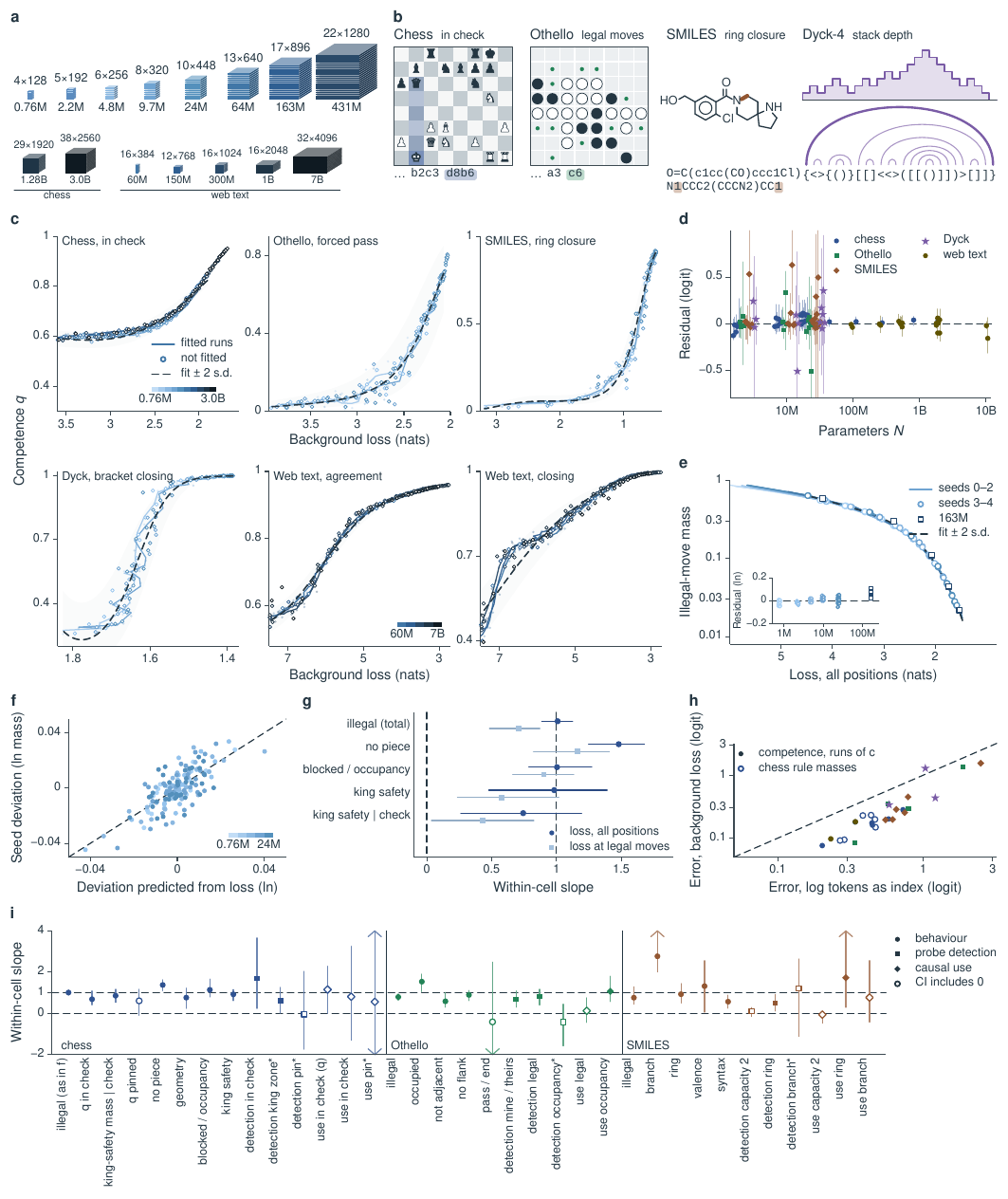}{ed1}{Fig.~B1}{Background loss indexes competence across domains, sizes and seeds.}{\textbf{a}, Model sizes (layers $\times$ width) to scale: 0.76M--431M (top); chess 1.28B, 3.0B and web text 60M--7B at one quarter scale (bottom). \textbf{b}, Example sequences, skill variable highlighted. \textbf{c}, Competence $q$ against background loss for every fully supervised run: chess in check (9.7M--3.0B, 11 runs); Othello, SMILES and Dyck (2.2M--24M, two seeds, 6 runs each); web-text agreement and closing (60M--7B, two recipes, 12 and 13 runs). Dashed, polynomial in background loss fitted to logit $q$ of the smaller seed-0 models (lines); shading, $\pm2$ residual s.d.; open circles, runs not fitted (other seeds, larger models); colour, model size. \textbf{d}, Residual of every run not fitted, for the skills of \textbf{c} and 15 more (98 runs; marker, median over snapshots; bar, interquartile range). \textbf{e}, Chess illegal-move mass against loss on all positions: seeds 0--2 at 0.76M--64M fitted, seeds 3--4 and 163M not fitted; inset, their residuals against $N$. \textbf{f}, Within-cell deviation of ln illegal mass against the deviation predicted from each checkpoint's own loss (160 checkpoints, 32 cells differing only in seed). \textbf{g}, Within-cell slopes for five rule masses indexed by loss on all positions or at legal moves. \textbf{h}, Fit error with background loss or log training tokens as the index (21 skills, domain as in \textbf{d}; open circles, six chess rule masses, one size left out). \textbf{i}, Within-cell slopes for behaviour, probe-detection and causal-use metrics in chess, Othello and SMILES (66, 66 and 65 cells; 32 for the illegal mass of \textbf{f}); asterisks, incidental variables; open symbols, CI includes 0; arrows, clipped intervals. Dashed lines, $y=x$, zero, slope 1. Error bars, 95\% cell-bootstrap CIs. Unit: snapshot (\textbf{c}, \textbf{e}), run (\textbf{d}), cell (\textbf{f}, \textbf{g}, \textbf{i}).}

\ppappfigure{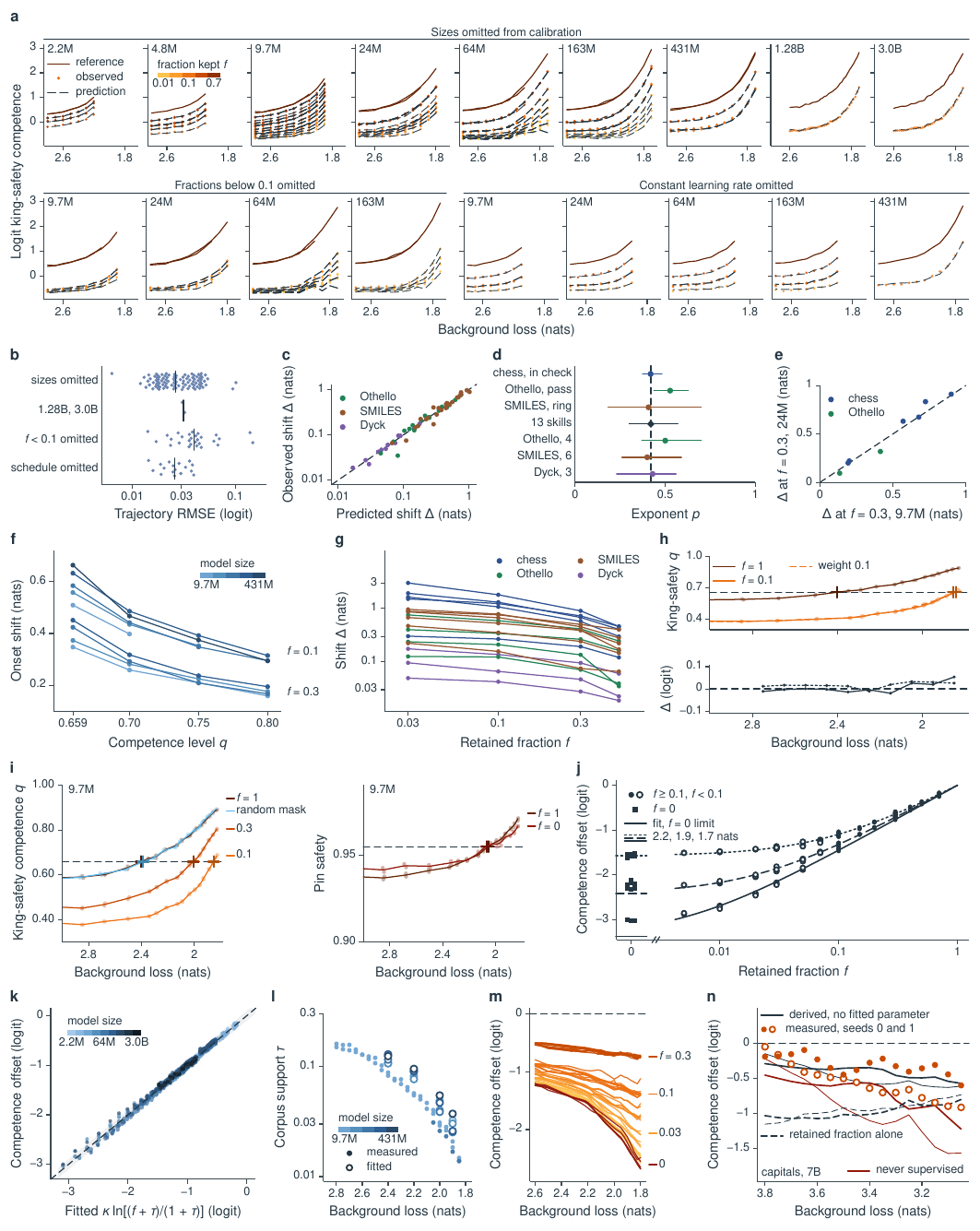}{ed2}{Fig.~B2}{One relation gives the shift for held-out fractions, sizes and skills.}{\textbf{a}, The 146 chess test trajectories: logit king-safety competence against background loss for sizes, fractions below 0.1 and the constant learning rate omitted from calibration. Lines, fully supervised; dots, observed (colour, $f$); dashed, prediction. \textbf{b}, Root-mean-square error per test trajectory (bar, median; $n=99$, 2, 28, 17). \textbf{c}, Leave-one-skill-out predicted against observed shift for 13 Othello, SMILES and Dyck skills at 9.7M (52 cells). \textbf{d}, Exponent $p$ of the shift form per skill and domain (bars, 95\% bootstrap intervals; dashed, pooled). \textbf{e}, Shift at $f=0.3$ at 24M against 9.7M. \textbf{f}, Onset shift at four competence levels, constant-learning-rate models, 9.7M--431M. \textbf{g}, Shift of 18 skills against $f$ at 9.7M (colour, domain). \textbf{h}, Every in-check target at loss weight 0.1 (dashed) against 10\% kept at full weight (solid), 9.7M, two seeds; below, their difference (logit). \textbf{i}, King-safety competence after thinning in-check targets or masking as many random targets (left), and pin safety without in-check targets (right), 9.7M (two seeds; $+$, onset). \textbf{j}, Logit offsets against $f$ at 2.2, 1.9 and 1.7 nats, with the effective-supervision fit, its $f=0$ limit and the $f=0$ runs. \textbf{k}, Pooled fitted form: 4,312 offsets from 109 thinned chess trajectories (2.2M--3.0B) against $\kappa\ln[(f+\tau)/(1+\tau)]$, with $\kappa$ and $\tau$ fitted at each background loss (colour, model size; shading, $\pm0.1$ logit). \textbf{l}, Corpus support $\tau$ by model size: filled, measured (9.7M, two seeds; 64M); open, fitted to the pooled offsets per size (9.7M--431M). \textbf{m}, Offsets of 48 thinned runs covering 1.8--2.6 nats (colour, $f$; red, $f=0$). \textbf{n}, Capitals at 7B ($f=0.3$; 16 background losses, two seeds): measured (points), derived with no fitted parameter (solid), retained fraction alone (dashed), never supervised (red); open and thin, seed 1. Dashed diagonals, equality. Units: trajectories (\textbf{a}, \textbf{b}), skill and fraction (\textbf{c}, \textbf{g}), otherwise runs.}

\ppappfigure{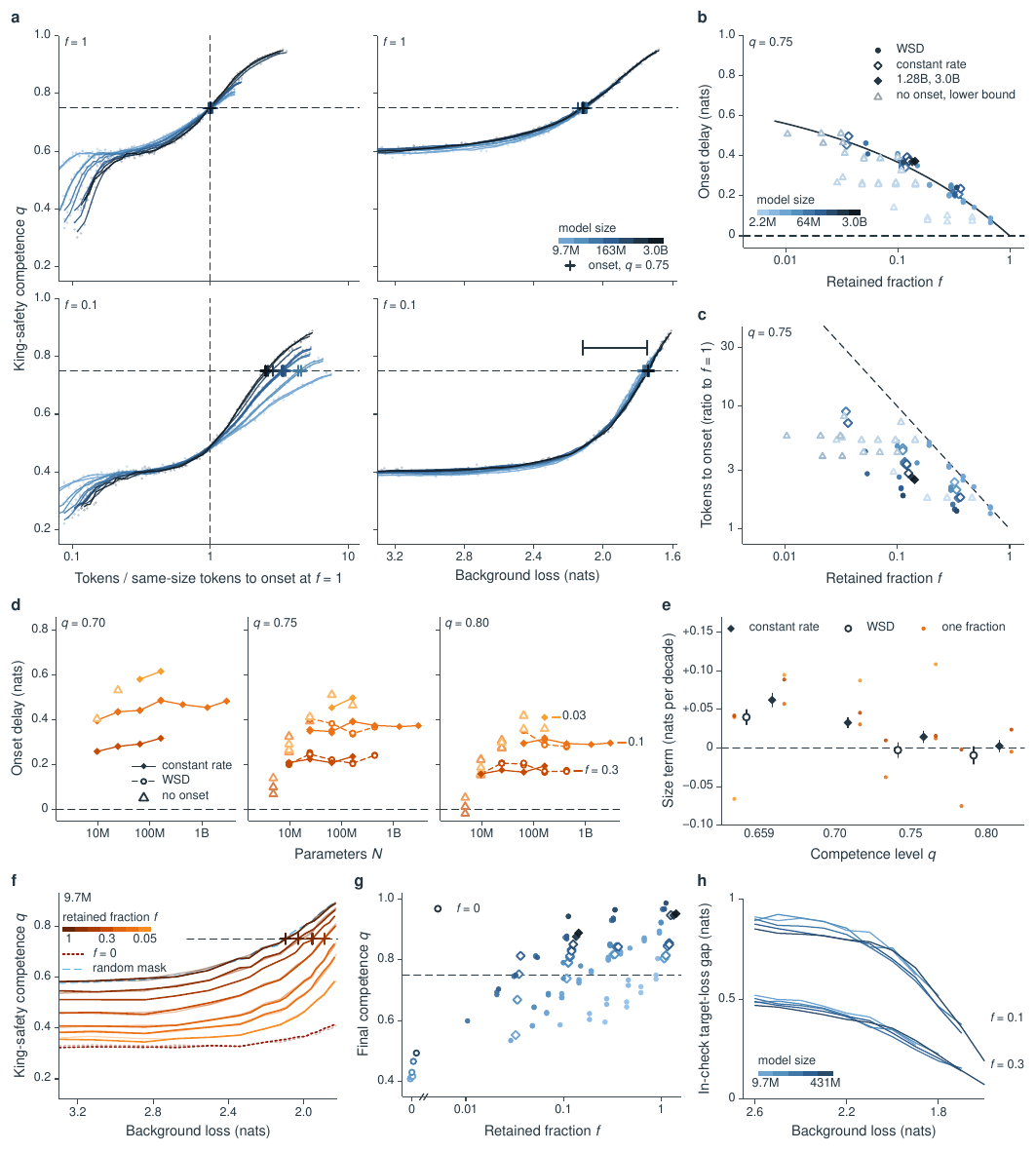}{ed3}{Fig.~B3}{Thinning delays chess king safety by the same amount from 9.7M to 3.0B parameters.}{Two schedules: warm-up--stable--decay (WSD), 2.2M--431M, and constant learning rate, 9.7M--3.0B. Delays are the onset of the fully supervised model of the same size and schedule minus the run's onset; light triangles, runs without onset at their lower bound. \textbf{a}, Competence $q$ against tokens relative to the same-size fully supervised onset (left) and against background loss (right), $f=1$ (top) and $f=0.1$ (bottom), 22 constant-rate runs, 9.7M--3.0B (colour, size; dots, snapshots; $+$, onsets at $q=0.75$; bracket, mean onsets of the two fractions). \textbf{b}, Onset delay at $q=0.75$ against $f$ for 85 thinned runs (50 with onset; circles, WSD; open diamonds, constant rate; filled, 1.28B and 3.0B). Curve, pooled fit $0.77\,(1-f^{0.28})$ over all sizes and both schedules. \textbf{c}, Tokens to onset divided by those at $f=1$, same runs; dashed, $1/f$. \textbf{d}, Onset delay against parameters $N$ at three competence levels for $f=0.3$, 0.1 and 0.03 (solid, constant rate, as Fig.~1e; dashed, WSD; lines, size means). \textbf{e}, Size term: slope of the delay against $\log_{10}N$ with one intercept per fraction (bars, $\pm1$ s.e.; small points, single fractions). Constant rate at $q=0.659$, 0.70, 0.75 and 0.80 (18, 17, 15, 12 runs); WSD at 0.659, 0.75 and 0.80 (30, 21, 18 runs). \textbf{f}, King-safety competence at 9.7M, eight fractions from 1 to 0, seeds 0--2 (seed 1 opaque), random-mask control dashed, $f=0$ dotted; $+$, seed-1 onsets. \textbf{g}, Final competence against $f$ for 120 runs, 2.2M--3.0B (markers as in b; open circles left of the break, $f=0$). \textbf{h}, In-check target loss of thinned minus fully supervised runs at matched background loss, $f=0.1$ and 0.3, 9.7M--431M. Unit: one training run.}

\ppappfigure{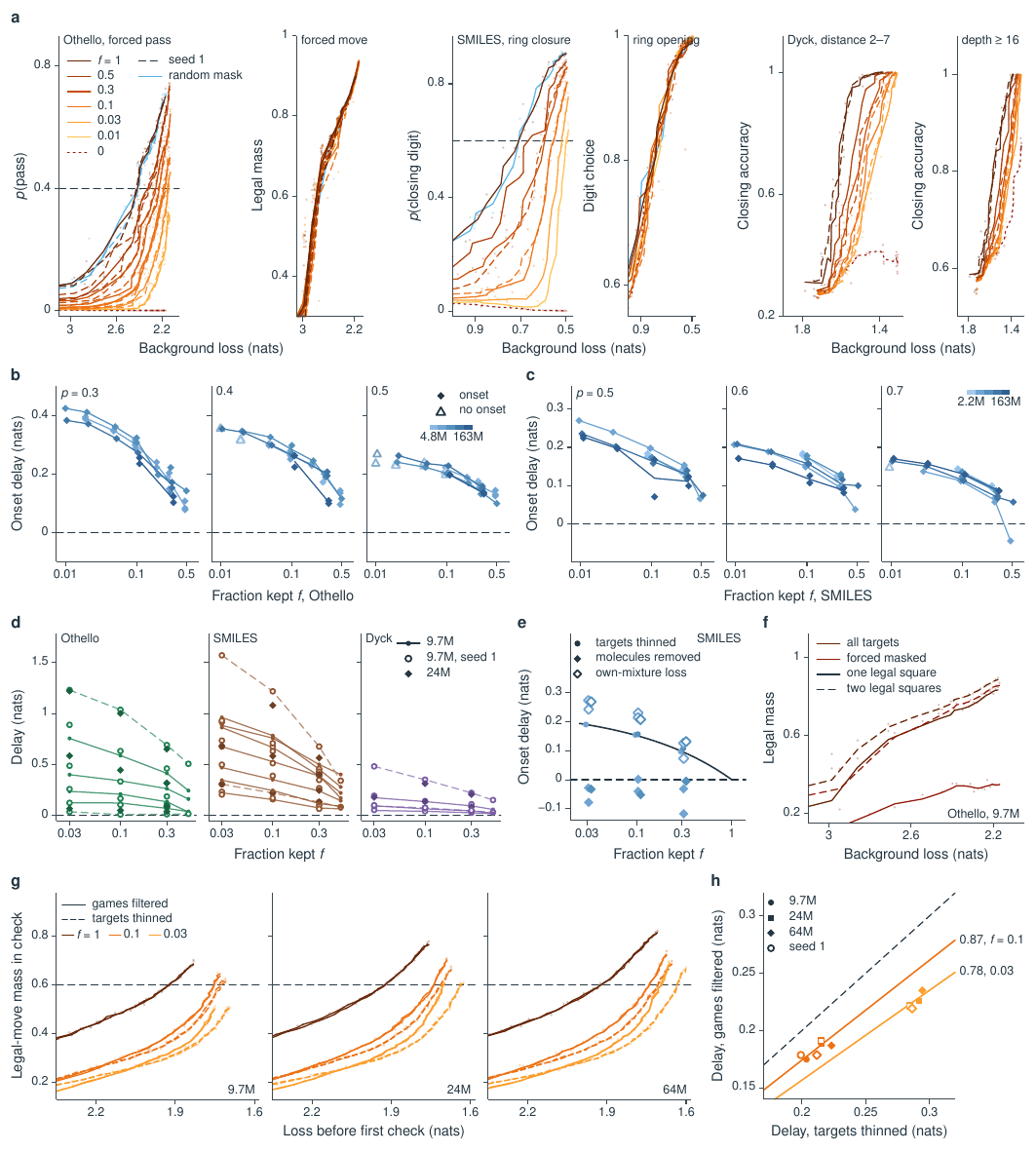}{ed4}{Fig.~B4}{The delay holds in Othello, SMILES and Dyck, and for filtered games.}{\textbf{a}, Competence of one thinned skill per domain against background loss at 9.7M (left of each pair), with an untouched skill of the same runs (right): Othello forced pass and forced-move legality (21 runs), SMILES ring closure and ring opening (10 runs), Dyck closes at match distance 2--7 and at depth $\geq16$ (10 runs). Colour, $f$; dashed, seed 1; blue, random mask; lines, five-evaluation smoothing in logit; dashed horizontal, onset level. \textbf{b}, Othello forced-pass onset delay relative to the same-size fully supervised model at $p(\text{pass})=0.3$, 0.4 and 0.5 for 35 thinned runs, 4.8M--163M (colour, size; lines, size means; triangles, lower bounds for runs without onset). \textbf{c}, The same for SMILES ring closure at 0.5, 0.6 and 0.7, 30 runs, 2.2M--163M. \textbf{d}, Delay, the mean horizontal distance from the fully supervised trajectory of the same size and seed over 20--80\% of the skill's logit range, for every measured skill in Othello (6), SMILES (9) and Dyck (5): 151 runs; lines, one skill at 9.7M. \textbf{e}, SMILES ring closure under target thinning (circles; curve, 9.7M masking curve) and under removal of ring-containing molecules to the same fraction (diamonds), relative to the unthinned onset; filled, on the shared background loss; open, on the loss of each run's own training mixture. Colour, size as in \textbf{b}. \textbf{f}, Othello legal mass on positions with one and with two legal squares when forced-move targets are masked. \textbf{g}, Chess legal-move mass in check for the 30 runs of Fig.~1f, one panel per size. \textbf{h}, Delay of game filtering against delay of target thinning at the same residual share, legal mass 0.6 (one point per size and seed). Lines, fitted ratios; dashed, equal delay. Unit: one training run.}

\ppappfigure{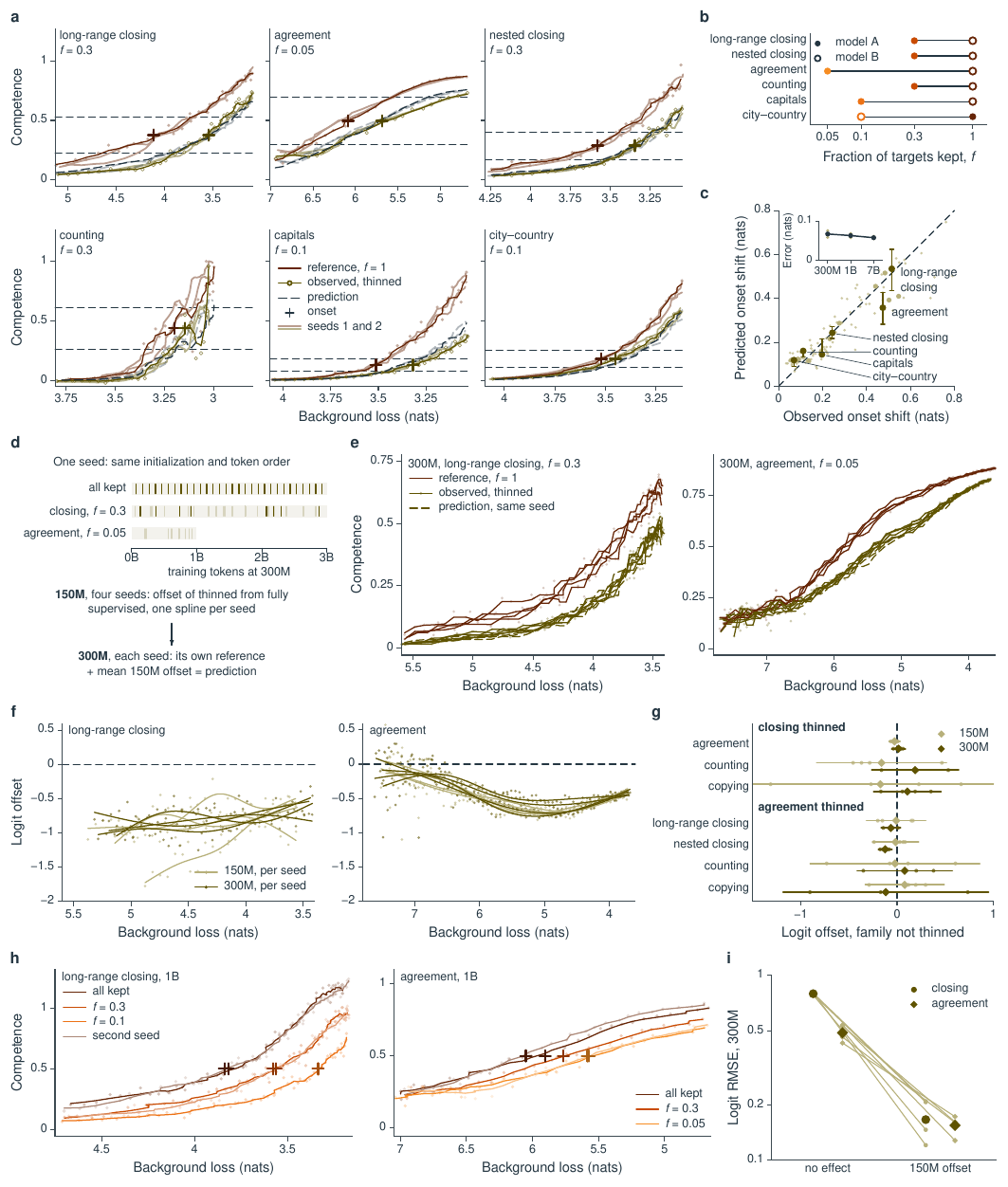}{ed5}{Fig.~B5}{In web text, thinning one skill shifts only that skill, by an offset that transfers across size.}{\textbf{a}, All six 7B skill contrasts: competence against background loss for the fully supervised model (reference), its trajectory shifted by the 60M--1B response (prediction) and the thinned model (observed), seed 0 (dots, checkpoints; lines, running medians of three followed by a three-point mean; $+$, onsets at competence 0.5; dashed, 0.3 and 0.7) with seeds 1 and 2 behind (faint lines). \textbf{b}, Fraction of each skill's targets kept in the two 7B models. \textbf{c}, Predicted against observed onset shift at 7B (dark, seed 0 means over five competence levels, bars, range over calibration sizes; light, seeds 1 and 2; small, single levels; dashed, $y=x$). Inset, mean absolute onset error of the same two designs at 300M, 1B (two seeds each) and 7B (three seeds; dots, seeds). \textbf{d}, Isolated thinning: per seed, three runs differ only in one family's mask. The offset (logit competence, thinned minus fully supervised, at matched background loss) is fitted per 150M seed, averaged and added to each 300M seed's reference. \textbf{e}, Each 300M seed and its prediction. \textbf{f}, Offsets at both sizes (dots, grid points; lines, one spline per seed). \textbf{g}, Offsets of the families that were not thinned (diamonds, means; bars, 95\% intervals). \textbf{h}, 1B trajectories at every fraction ($+$, onsets at 0.5). \textbf{i}, Logit RMSE of each 300M seed's trajectory without an effect and with the transferred offset (large, geometric means). Four seeds per size in \textbf{d}--\textbf{g}, \textbf{i}; intervals, 95\% $t$ over seeds; three seeds per 7B model. Unit: one run.}

\clearpage
\section{Outputs of a never-supervised model}\label{app:C}\setcounter{equation}{0}

\subsection{Output classes at the skill's positions}
The eight classes are: legal king move; legal capture of the checking piece, by the king or by another piece; legal interposition; move of another piece that obeys the piece-movement rules and leaves the king attacked only by the checking piece, which would be legal if the king were not in check; king move to an attacked square; move that exposes the king to a further piece; token that is not a move of the position; and end-of-game and padding tokens. The legal classes coincide with the legal moves at every position. Probabilities are means over positions with 95\% intervals from 1,000 bootstrap resamples of games.

At its last snapshot, at a background loss of 1.62 nats, the 64M model trained without in-check targets placed 0.48 (0.47--0.49) of its probability on moves that ignore the check. The share was 0.46--0.50 at every background loss below 2.2 nats, in both sizes and both 9.7M seeds. Among the legal replies, its most probable reply was the move played at 0.636 of the 7,149 positions with two or more legal replies, against 0.676 for the fully supervised model and 0.304 by chance; at 9.7M the values were 0.591 and 0.594 for the two seeds against 0.619--0.636 for three fully supervised seeds.

The returns of Fig.~2c start from an early and a late snapshot of each source run, with the warm-up before the return. In the return from the last 64M snapshot, the probability on moves that ignore the check fell from 0.48 to 0.14 within eight steps, the probability on legal king moves rose from 0.20 to 0.47 and the legal-move mass from 0.44 to 0.73. The gain in legal-move mass over the first eight steps was 0.18 and 0.24 at 9.7M, from snapshots at 1.99 and 1.83 nats, and 0.23 and 0.30 at 64M, from snapshots at 1.95 and 1.62 nats.

At the last snapshot of the 9.7M SMILES model trained without ring-closure targets, at 0.498 nats, 0.735 of the probability lay on an atom, 0.228 on opening a branch and 0.0004 on the true closing digit. After all ring-closure targets returned to that snapshot, at 16 sequences per step, the true digit had a probability of 0.043 after one step, 0.36 after two and 0.60 after 12.

\subsection{Chess games without a check}
\textit{Stream and positions.} The stream keeps, in order, every game of the chess training stream in which a replay finds no position with the side to move in check: 9.1\% of games, with a mean length of 43 moves against 70 for games with a check. Each evaluation prompt runs from the separator to the checking move of the first check of a held-out game. The reference checkpoints are within 0.03 nats of the model trained on games without a check in held-out loss before the first check. The ranking by move frequency counts moves by side to move in the first 600M tokens of the stream. Table~C1 gives every measure for the four sizes.

\textit{Moves of the same piece.} At each first check, a legal reply is paired with a move from the same square that obeys the piece-movement rules and leaves the king attacked, and the score is the share of pairs in which the legal reply has the higher logit, with ties counted as one half, averaged over the pairs of a game and then over games. With both moves restricted to moves of a piece other than the king that neither capture, promote nor give check, the legal reply is an interposition (847 games, 9,988 pairs). The 9.7M model trained on games without a check preferred it in 0.772 of comparisons (95\% interval 0.757--0.787), against 0.561 for the ranking by move frequency.

\textit{Board pairs.} One prompt is kept for each distinct position (1,681; piece placement, side to move, castling and en-passant rights), the shortest and then the first in the set. For two moves $A$ and $B$ of a pair, the shift measure asks whether the logit of $A$ minus that of $B$ is larger on the board where $A$ is legal than on the board where $B$ is legal; a ranking that does not depend on the board scores one half on it and zero on the share of pairs with the legal move preferred on both boards. Several pairs of moves on the same two boards are averaged. Intervals resample the 1,162 contributing games 2,000 times, and a board pair is weighted by the product of the counts of its two games.

\textit{Transfer between replies (Fig.~C1).} Each training example is a prefix of a game outside the evaluation set that ends at the first check of the game, with one loss, on the reply played. The three sets of replies are checks by a single knight, checks by a single bishop, rook or queen whose every legal reply interposes a piece, and first checks of any kind; the two controls are the moves two plies earlier in the knight games, with the same side to move, and the knight replies shuffled among positions with the same side to move. Each run takes 96 updates of 32 prefixes with a new AdamW state, a learning rate of $4\times10^{-4}$ after a ten-update warm-up, no weight decay and gradient clipping at 1. First checks of the training prefixes fall between plies 4 and 159, and positions shared with the evaluation set are excluded. The five sets hold 3,072 prefixes from distinct games and differ in prefix length. Intervals resample the evaluation games 10,000 times, with all runs and both evaluations of a game kept together, and describe one run per set. At the 92 positions where every legal reply blocks, legal-move mass changed by $-0.040$ ($-0.060$ to $-0.020$) after the replies to knight checks, by the same amount after the ordinary moves (difference $+0.000$, $-0.030$ to $+0.031$), by $+0.235$ (0.181--0.289) after first checks of any kind and by $+0.506$ (0.449--0.564) after forced blocks. At the 326 knight checks it was 0.569 before and 0.897, 0.551, 0.861 and 0.616 after the knight replies, the forced blocks, the first checks of any kind and the ordinary moves. Knight replies shuffled among positions lowered it to 0.139 at knight checks and 0.019 at forced blocks. At the 860 positions where a block is among the legal replies, the replies to knight checks raised legal-move mass from 0.443 to 0.583.

\pptablecaption{tab:s6}{Table~C1}{First checks of held-out games by model size.}{Chess models of seed 1 trained on games without a check, without in-check targets and fully supervised, at the first check of 1,704 held-out games. Tokens, training tokens of the checkpoint (millions); loss, held-out loss before the first check (nats); legal, legal-move mass. Ranked first, share of the 1,533 positions with two or more legal replies whose most probable legal reply is the move played (uniform choice 0.290, move frequency 0.405). Quiet block, share of comparisons, averaged by game (847 games), in which a legal interposition has a higher logit than an unsafe move of the same piece, neither a capture, promotion or check. Shift and both boards, the two measures of the 18,701 board pairs, with the 95\% interval over games for the second; disjoint, both boards with each game in at most one pair (398 pairs).}
\begin{pptable}{llrrrrrrrr}{0}
\toprule
\pptablehead{Size} & \pptablehead{Trained} & \pptablehead{Tokens} & \pptablehead{Loss} & \pptablehead{Legal} & \pptablehead{Ranked first} & \pptablehead{Quiet block} & \pptablehead{Shift} & \pptablehead{Both boards} & \pptablehead{Disjoint} \\
\midrule
\endhead
9.7M & Games without a check & 600 & 1.803 & 0.453 & 0.652 & 0.772 & 0.905 & 0.646 (0.601--0.687) & 0.566 \\
 & Without in-check targets & 302 & 1.778 & 0.426 & 0.672 & 0.666 & 0.882 & 0.624 (0.584--0.658) & 0.571 \\
 & Fully supervised & 200 & 1.831 & 0.847 & 0.678 & 0.963 & 0.996 & 0.933 (0.908--0.956) & 0.927 \\
\addlinespace[2pt]
24M & Games without a check & 600 & 1.747 & 0.477 & 0.661 & 0.796 & 0.928 & 0.697 (0.657--0.736) & 0.656 \\
 & Without in-check targets & 200 & 1.760 & 0.437 & 0.645 & 0.682 & 0.886 & 0.642 (0.603--0.678) & 0.592 \\
 & Fully supervised & 200 & 1.763 & 0.887 & 0.710 & 0.973 & 0.998 & 0.960 (0.943--0.973) & 0.947 \\
\addlinespace[2pt]
64M & Games without a check & 600 & 1.713 & 0.507 & 0.666 & 0.823 & 0.952 & 0.749 (0.712--0.785) & 0.699 \\
 & Without in-check targets & 200 & 1.713 & 0.447 & 0.676 & 0.709 & 0.914 & 0.710 (0.675--0.742) & 0.647 \\
 & Fully supervised & 200 & 1.708 & 0.908 & 0.711 & 0.975 & 0.999 & 0.972 (0.960--0.983) & 0.963 \\
\addlinespace[2pt]
163M & Games without a check & 450 & 1.698 & 0.518 & 0.676 & 0.825 & 0.953 & 0.775 (0.741--0.808) & 0.715 \\
 & Without in-check targets & 162 & 1.715 & 0.435 & 0.671 & 0.694 & 0.905 & 0.668 (0.630--0.703) & 0.600 \\
 & Fully supervised & 150 & 1.718 & 0.903 & 0.703 & 0.973 & 0.998 & 0.969 (0.956--0.980) & 0.948 \\
\bottomrule
\end{pptable}

\clearpage
\ppappfigure{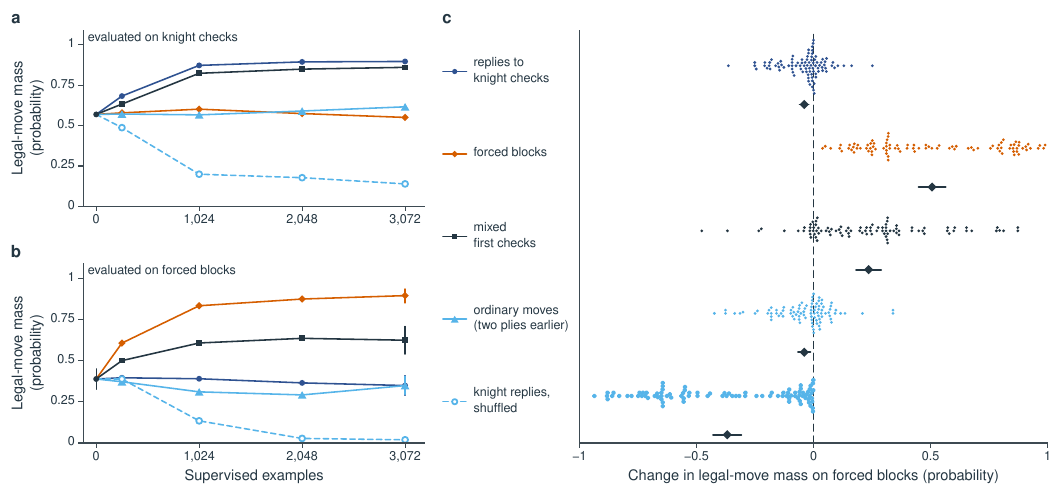}{s11}{Fig.~C1}{Transfer between two kinds of reply to check.}{The 9.7M chess model trained on games without a check received 3,072 supervised examples of one kind. \textbf{a}, Legal-move mass at 326 held-out first checks by a knight. \textbf{b}, The same at 92 held-out first checks whose every legal reply interposes a piece (forced blocks); bars, 95\% intervals at start and end. \textbf{c}, Change in legal-move mass in each of these 92 games; diamonds and bars, means and paired 95\% bootstrap intervals. Lines in \textbf{a} and \textbf{b} as named in \textbf{c}; mixed first checks, first checks of any kind; ordinary moves, the moves two plies before the first check of the knight games; shuffled, knight replies permuted among positions. One run per kind of example.}

\clearpage
\section{Acquisition cost and the cost relation}\label{app:D}\setcounter{equation}{0}

\subsection{Acquisition cost after targets return}
Stored checkpoints hold the weights only, so the optimizer state is rebuilt at every return. Costs count the tokens after the targets return. The steps that rebuild the optimizer state contain none of the returned targets and are not charged; charging them gives 3.1--7.5\% for the chess source runs at their lowest-loss checkpoints (Fig.~D1e). With the warm-up at the return, no parameter update precedes the returned targets (Fig.~D1b).

\textit{Chess.} Returns use the peak learning rate of the main recipe, 0.004. A run stops after two consecutive evaluations with king-safety competence $q$ at or above 0.80 for $s=1$ and 0.70 for smaller shares, or at its token budget. Budgets grow as the returned share falls: 40M tokens at $s=1$, 60M at 0.3, 80M--200M at 0.1 and 100M--200M at 0.03; two returns at $s=0.3$ and 0.1 ran with the 40M budget. The 198 chess returns of the cost relation are the 110 returns with the warm-up before the return, the 82 returns with the warm-up at the return whose checkpoint and share have no return with the warm-up before it, and six returns of all targets from 9.7M and 64M checkpoints at 2.99--3.99 nats, which use the warm-up before the return and a budget of 150M tokens. The two chess scores cross their levels together: at the first crossing of $q=0.659$ in a return, the legal-move mass has a median of 0.60, with 10th and 90th percentiles of 0.56 and 0.61 across 326 runs.

\textit{Web text.} The eleven source models are seed 0 at 9.7M, 24M, 64M and 163M, seeds 1 and 2 at 9.7M--64M and seed 1 at 163M. After the moments are estimated the first moment is set to zero, and the 30 steps on the source model's own targets run at the learning rate of the source schedule, 0.008. Targets return from 7--9 checkpoints of each source run. Evaluations fall at steps 0--6, 8, 10, 12, 14 and 16 and then at steps 15\% apart. A run stops once competence is at 0.90, or at twice the steps of its first evaluation at or above 0.85 with at least 24 steps, or at its budget, which is 60M tokens for returns near or above the acquisition loss and 30M otherwise; every run reached the level.

\textit{SMILES.} Returns continue the data order of the source run at a learning rate of 0.004. Evaluations fall when the targets return and then at steps spaced by a factor of 1.25. A run stops after two consecutive evaluations at 0.75 or above (0.7 at shares below 1) or at a budget of 20M, 30M, 60M and 100M tokens for shares of 1, 0.3, 0.1 and 0.03.

\textit{Othello.} The eight source runs are four at 9.7M (two seeds trained with none of the forced-pass targets, and one each with 10\% and 30\%), two at 24M (none and 10\%) and two at 64M (10\% and 30\%), and their checkpoints lie at background losses of 2.12--2.86 nats. Returns use a learning rate of 0.004, with the moments estimated over 20 steps without parameter updates, and are evaluated on the chess schedule. A run stops after two consecutive evaluations at 0.6 or above. Eleven runs were at the level when the targets returned and are counted as one step, and three more reached it within the first step; these fourteen are upper bounds.

\subsection{Logit distance and the cost relation}
The reference has a seed other than that of the source run for 50 chess, 21 web-text and 2 Othello runs. References cover 186 of 198 chess runs, 88 of 90 web-text runs, 66 of 73 SMILES runs and 50 of 53 Othello runs. For the progress term and the split at $L^*(s)$, the acquisition loss is read from one curve per domain, fitted across sizes and returned shares in chess, across returned shares at 9.7M in SMILES and across sizes in web text; it lies within 0.04 nats of the value from the same-size reference in chess and SMILES and within 0.08 nats in web text.

Table~D1 gives the fits. In SMILES the rate is $0.74\pm0.21$, $0.57\pm0.10$, $0.47\pm0.08$ and $0.52\pm0.09$ decades per logit at 9.7M, 24M, 64M and 163M. The 13 returns of all ring-closure targets from a logit distance below $-1.3$ reached the level within 14--20 steps, inside the 20-step warm-up, at every size; without the returns of 25 steps or fewer the rate is $0.56\pm0.18$. For the 56 runs returned above the acquisition loss with a positive logit distance that reached the level in chess, web text and SMILES, the background loss at which they reached the acquisition level differed from $L^*(s)$ by a median of 0.04 nats in chess and web text and 0.01 nats in SMILES (Fig.~D2h), and the two-part form with the single line gave their costs with a median absolute error of 0.17 dex. Across seven acquisition levels in chess the rate is 0.78--1.20 decades per logit (Fig.~D2b). The standard errors of the single line and of the trend with model size are clustered by source run. One never-supervised Othello return, from a checkpoint 1.5 logits above the acquisition loss, did not reach the level within 306 steps. In the near-floor ordering a higher score counts as a lower predicted cost.

\subsection{Batch size and learning rate at the return}
The three checkpoints of each size are the first stored checkpoint at or below the acquisition loss, the lowest-loss checkpoint and the one nearest the midpoint of the two in background loss. Every setting is evaluated at every step up to step 40. The runs from initialization have two seeds at 9.7M and 24M and one at 64M. Table~D2 gives every cost.

Training from initialization needed 0.55--0.66 times the tokens at 16 sequences and 2.2--3.3 times at 256. At the standard learning rate, the five returns of 25--216 optimizer steps needed 1.1--1.65 times the tokens at 16 sequences, in 4.3--6.6 times as many steps, and 1.5--2.2 times the tokens at 256 sequences. With 16 sequences at one third of the learning rate, the tokens were 0.71--1.07 times those of the standard setting at 2.8--4.3 times the steps. Returns that finish within ten optimizer steps needed fewer tokens at 16 sequences, 0.42--0.94 times, so the costs of 0.10--0.78\% from the lowest-loss checkpoints are upper estimates of the data required. From the lowest-loss checkpoints the cost was 0.15--1.35\% of the tokens from initialization at the same setting across the five settings. The slope of $\log_{10}$ relative cost against the logit distance of the checkpoint was 0.86--1.29 decades per logit at the four other settings, and the single line of Fig.~4b gave the relative costs at the five settings with mean errors of $-0.40$ to $-0.12$ dex. From initialization the permuted rule needed 1.03--1.13 times the tokens of the true rule at these settings.

\subsection{Transplanted weight changes}
Table~D3 gives the legal-move mass and the change in background loss for every change and recipient checkpoint. At 9.7M the Frobenius norm of the change due to the returned targets was 1.7\% of that of the final weights for the 619M-token checkpoint and 11\% for the 97M-token checkpoint; at 64M, relative to the weights of its own checkpoint, it was 0.8--0.9\% at 1.93B tokens, 1.5\% at 226M and 3.2--3.3\% at 161M. Legal-move mass and background loss were evaluated on the 1,000 held-out games used for the returns.

At 9.7M, across the checkpoints with a measured acquisition cost, nine in one seed and six in the other, the rank correlation between the recipient's cost and the mass gained was $-0.85$ and $-0.49$ for the change from the 193M-token checkpoint. The weight change of a return without the subtraction of its control run raised legal-move mass at other checkpoints similarly but raised background loss by 0.16--0.26 nats for the changes from 619M, 290M and 193M tokens and by 0.32--0.44 for the change from 97M, as did the change of the control run alone; the cosine similarity between the two was 0.86--0.98. Changes computed at different checkpoints of a run had a cosine similarity of 0.06--0.23 with one another and of 0.00 between seeds. At 64M the Gaussian change gained between $-0.010$ and $-0.001$ and the change added to the other seed's checkpoints between $-0.000$ and $+0.002$ in legal-move mass.

\pptablecaption{tab:s2}{Table~D1}{Fits of the cost relation by domain.}{Fits of $\log_{10}$ acquisition cost to $\log_{10}[P+10^{\,a+b\min(x,0)}]$ over the 312 runs that reached the acquisition level, with logit distance $x$ and progress term $P$; rates in decades per logit, residual standard deviations in dex. The rates of the main text are from the maximum-likelihood fit with an intercept and a rate for each domain, a random intercept for each source run (standard deviation 0.11 dex) and a residual variance for each domain; the rates differ between domains (likelihood ratio $\chi^2=44.4$, 2 degrees of freedom). The least-squares columns fit each domain alone, and the single line fits one intercept and one rate to all runs. Values after $\pm$ are standard errors; the least-squares fits of single domains give the larger of the ordinary and the cluster-robust one, and the single line the cluster-robust one.}
\begin{pptable}[small]{lrrrrrrrrr}{0}
\toprule
 & & & & \multicolumn{3}{c}{\pptablehead{Fit with source-run intercepts}} & \multicolumn{3}{c}{\pptablehead{Least squares}} \\
\cmidrule(lr){5-7}\cmidrule(lr){8-10}
\pptablehead{Domain} & \pptablehead{Runs} & \pptablehead{Source} & \pptablehead{Below acq.} & \pptablehead{Intercept $a$} & \pptablehead{Rate $b$} & \pptablehead{Residual} & \pptablehead{Intercept $a$} & \pptablehead{Rate $b$} & \pptablehead{Residual} \\
 &  & \pptablehead{runs} & \pptablehead{loss} &  &  & \pptablehead{s.d.} &  &  & \pptablehead{s.d.} \\
\midrule
\endhead
Chess & 159 & 12 & 129 & $-$0.53 $\pm$ 0.05 & 1.12 $\pm$ 0.04 & 0.25 & $-$0.52 & 1.14 $\pm$ 0.04 & 0.25 \\
Web text & 88 & 11 & 70 & $-$0.62 $\pm$ 0.05 & 0.91 $\pm$ 0.04 & 0.10 & $-$0.66 & 0.89 $\pm$ 0.06 & 0.17 \\
SMILES & 65 & 4 & 51 & $-$0.42 $\pm$ 0.09 & 0.59 $\pm$ 0.07 & 0.32 & $-$0.41 & 0.61 $\pm$ 0.09 & 0.37 \\
\addlinespace[2pt]
Single line & 312 & 27 & 250 &  &  &  & $-$0.53 & 0.97 $\pm$ 0.07 & 0.36 \\
\bottomrule
\end{pptable}

\pptablecaption{tab:s3}{Table~D2}{Returns at other batch sizes and learning rates.}{Returns of all in-check targets to three checkpoints of chess source models at each of three sizes, at five settings of batch size (sequences per step) and learning rate; 64 sequences at the standard learning rate is the standard setting. Values are geometric means of two data orders. Background loss and logit distance are those of the stored checkpoint. The denominator of each cost is the run from initialization at the same setting.}
\begin{pptable}[small]{llrrrrrrr}{0}
\toprule
 & & & & \multicolumn{3}{c}{\pptablehead{Standard learning rate}} & \multicolumn{2}{c}{\pptablehead{One third of it}} \\
\cmidrule(lr){5-7}\cmidrule(lr){8-9}
\pptablehead{Size} & \pptablehead{Checkpoint} & \pptablehead{Background} & \pptablehead{Logit} & \pptablehead{16 seq.} & \pptablehead{64 seq.} & \pptablehead{256 seq.} & \pptablehead{64 seq.} & \pptablehead{16 seq.} \\
 & \pptablehead{(tokens)} & \pptablehead{loss (nats)} & \pptablehead{distance} &  &  &  &  &  \\
\midrule
\endhead
\addlinespace[3pt]
\multicolumn{9}{l}{\pptablehead{Cost, \% of the tokens from initialization at the same setting}} \\
\addlinespace[1pt]
9.7M & 77M & 2.21 & $-$0.04 & 48.1 & 24.0 & 11.3 & 22.7 & 30.2 \\
 & 145M & 2.04 & $-$0.40 & 31.7 & 10.6 & 4.79 & 9.39 & 14.3 \\
 & 619M & 1.83 & $-$1.10 & 1.35 & 0.788 & 0.652 & 0.895 & 0.969 \\
24M & 85M & 2.17 & $-$0.15 & 22.0 & 11.3 & 6.50 & 11.5 & 13.3 \\
 & 146M & 1.96 & $-$0.66 & 6.31 & 3.52 & 2.48 & 3.83 & 3.59 \\
 & 390M & 1.74 & $-$1.37 & 0.555 & 0.664 & 0.507 & 0.626 & 0.512 \\
64M & 97M & 2.21 & $-$0.06 & 17.8 & 10.9 & 10.9 & 10.5 & 10.8 \\
 & 193M & 1.88 & $-$0.97 & 1.02 & 0.896 & 1.17 & 1.14 & 0.861 \\
 & 644M & 1.62 & $-$1.97 & 0.155 & 0.246 & 0.335 & 0.303 & 0.175 \\
\addlinespace[3pt]
\multicolumn{9}{l}{\pptablehead{Optimizer steps to the acquisition level}} \\
\addlinespace[1pt]
9.7M & 77M & 2.21 & $-$0.04 & 945 & 216 & 83.6 & 315 & 852 \\
 & 145M & 2.04 & $-$0.40 & 624 & 94.8 & 35.5 & 130 & 402 \\
 & 619M & 1.83 & $-$1.10 & 26.6 & 7.1 & 4.8 & 12.4 & 27.3 \\
24M & 85M & 2.17 & $-$0.15 & 409 & 80.4 & 33.6 & 129 & 345 \\
 & 146M & 1.96 & $-$0.66 & 117 & 25.0 & 12.8 & 42.9 & 93.0 \\
 & 390M & 1.74 & $-$1.37 & 10.3 & 4.7 & 2.6 & 7.0 & 13.3 \\
64M & 97M & 2.21 & $-$0.06 & 394 & 91.2 & 49.3 & 134 & 361 \\
 & 193M & 1.88 & $-$0.97 & 22.6 & 7.5 & 5.3 & 14.6 & 28.8 \\
 & 644M & 1.62 & $-$1.97 & 3.4 & 2.1 & 1.5 & 3.9 & 5.8 \\
\addlinespace[3pt]
\multicolumn{9}{l}{\pptablehead{Tokens to the acquisition level from initialization, millions}} \\
\addlinespace[1pt]
9.7M &  &  &  & 64 & 118 & 389 & 182 & 92 \\
24M &  &  &  & 61 & 93 & 271 & 147 & 85 \\
64M &  &  &  & 73 & 110 & 236 & 168 & 110 \\
\addlinespace[3pt]
\multicolumn{9}{l}{\pptablehead{Tokens of the permuted rule relative to the true rule, 9.7M}} \\
\addlinespace[1pt]
9.7M & 77M & 2.21 & $-$0.04 & 1.3 & 1.6 & 1.8 & 1.8 & 1.8 \\
 & 145M & 2.04 & $-$0.40 & 1.6 & 2.5 & 3.4 & 3.0 & 2.5 \\
 & 619M & 1.83 & $-$1.10 & 11.0 & 12.7 & 11.0 & 12.7 & 12.7 \\
\bottomrule
\end{pptable}

\pptablecaption{tab:s4}{Table~D3}{Transplanted weight changes in chess.}{In-check legal-move mass after the change due to the returned targets, computed at one checkpoint (column groups, by training tokens), is added without training to a checkpoint of the same source run (rows), with the change in background loss ($\Delta L$, nats). The last columns give the mass on the permuted rule's targets after the weight change of a permuted-rule return is added. Values are means of the two source runs; bold marks the checkpoint at which the change was computed. \textbf{a}, 9.7M models. \textbf{b}, 64M models.}
\begin{pptable}[small]{lrrrrrrrrrrrr}{0}
\toprule
\multicolumn{13}{l}{\pptablehead{a, 9.7M}} \\
\addlinespace[1pt]
 & & \multicolumn{2}{c}{\pptablehead{619M}} & \multicolumn{2}{c}{\pptablehead{290M}} & \multicolumn{2}{c}{\pptablehead{193M}} & \multicolumn{2}{c}{\pptablehead{97M}} & \multicolumn{3}{c}{\pptablehead{Permuted rule}} \\
\cmidrule(lr){3-4}\cmidrule(lr){5-6}\cmidrule(lr){7-8}\cmidrule(lr){9-10}\cmidrule(lr){11-13}
\pptablehead{Recipient} & \pptablehead{No change} & \pptablehead{Mass} & \pptablehead{$\Delta L$} & \pptablehead{Mass} & \pptablehead{$\Delta L$} & \pptablehead{Mass} & \pptablehead{$\Delta L$} & \pptablehead{Mass} & \pptablehead{$\Delta L$} & \pptablehead{619M} & \pptablehead{290M} & \pptablehead{193M} \\
\midrule
\endhead
97M & 0.29 & 0.35 & $+$0.01 & 0.40 & $+$0.03 & 0.45 & $+$0.04 & \textbf{0.55} & $+$0.06 & 0.01 & 0.06 & 0.16 \\
116M & 0.30 & 0.37 & $+$0.01 & 0.43 & $+$0.03 & 0.49 & $+$0.04 & 0.58 & $+$0.06 & 0.02 & 0.10 & 0.25 \\
145M & 0.31 & 0.40 & $+$0.02 & 0.47 & $+$0.03 & 0.54 & $+$0.05 & 0.61 & $+$0.07 & 0.03 & 0.19 & 0.39 \\
193M & 0.32 & 0.43 & $+$0.02 & 0.52 & $+$0.04 & \textbf{0.59} & $+$0.05 & 0.61 & $+$0.07 & 0.06 & 0.31 & \textbf{0.61} \\
242M & 0.33 & 0.46 & $+$0.02 & 0.56 & $+$0.04 & 0.61 & $+$0.05 & 0.61 & $+$0.07 & 0.11 & 0.43 & 0.45 \\
290M & 0.34 & 0.48 & $+$0.02 & \textbf{0.60} & $+$0.04 & 0.61 & $+$0.05 & 0.61 & $+$0.07 & 0.16 & \textbf{0.61} & 0.36 \\
387M & 0.36 & 0.54 & $+$0.03 & 0.62 & $+$0.04 & 0.62 & $+$0.05 & 0.61 & $+$0.07 & 0.30 & 0.43 & 0.23 \\
484M & 0.36 & 0.57 & $+$0.03 & 0.62 & $+$0.03 & 0.62 & $+$0.04 & 0.60 & $+$0.07 & 0.41 & 0.33 & 0.16 \\
619M & 0.37 & \textbf{0.61} & $+$0.03 & 0.62 & $+$0.03 & 0.61 & $+$0.04 & 0.59 & $+$0.07 & \textbf{0.61} & 0.22 & 0.10 \\
\bottomrule
\end{pptable}

\begin{pptable}[small]{lrrrrrrrrrr}{0}
\toprule
\multicolumn{11}{l}{\pptablehead{b, 64M}} \\
\addlinespace[1pt]
 & & \multicolumn{2}{c}{\pptablehead{1.93B}} & \multicolumn{2}{c}{\pptablehead{226M}} & \multicolumn{2}{c}{\pptablehead{161M}} & \multicolumn{3}{c}{\pptablehead{Permuted rule}} \\
\cmidrule(lr){3-4}\cmidrule(lr){5-6}\cmidrule(lr){7-8}\cmidrule(lr){9-11}
\pptablehead{Recipient} & \pptablehead{No change} & \pptablehead{Mass} & \pptablehead{$\Delta L$} & \pptablehead{Mass} & \pptablehead{$\Delta L$} & \pptablehead{Mass} & \pptablehead{$\Delta L$} & \pptablehead{1.93B} & \pptablehead{226M} & \pptablehead{161M} \\
\midrule
\endhead
97M & 0.28 & 0.31 & $+$0.02 & 0.37 & $+$0.03 & 0.39 & $+$0.23 & 0.01 & 0.03 & 0.08 \\
113M & 0.30 & 0.33 & $+$0.02 & 0.40 & $+$0.03 & 0.43 & $+$0.22 & 0.00 & 0.06 & 0.15 \\
129M & 0.31 & 0.34 & $+$0.01 & 0.43 & $+$0.03 & 0.47 & $+$0.21 & 0.00 & 0.10 & 0.24 \\
161M & 0.33 & 0.38 & $+$0.01 & 0.50 & $+$0.03 & \textbf{0.57} & $+$0.15 & 0.01 & 0.21 & \textbf{0.61} \\
193M & 0.36 & 0.41 & $+$0.01 & 0.55 & $+$0.03 & 0.57 & $+$0.09 & 0.01 & 0.34 & 0.29 \\
226M & 0.37 & 0.42 & $+$0.01 & \textbf{0.60} & $+$0.03 & 0.57 & $+$0.07 & 0.01 & \textbf{0.62} & 0.19 \\
258M & 0.37 & 0.43 & $+$0.01 & 0.58 & $+$0.02 & 0.55 & $+$0.05 & 0.01 & 0.37 & 0.12 \\
322M & 0.40 & 0.48 & $+$0.01 & 0.63 & $+$0.02 & 0.60 & $+$0.03 & 0.02 & 0.37 & 0.11 \\
387M & 0.41 & 0.49 & $+$0.01 & 0.63 & $+$0.02 & 0.59 & $+$0.03 & 0.03 & 0.31 & 0.08 \\
516M & 0.43 & 0.53 & $+$0.02 & 0.63 & $+$0.02 & 0.60 & $+$0.02 & 0.07 & 0.22 & 0.05 \\
644M & 0.44 & 0.55 & $+$0.02 & 0.65 & $+$0.02 & 0.61 & $+$0.02 & 0.13 & 0.20 & 0.05 \\
773M & 0.45 & 0.57 & $+$0.02 & 0.64 & $+$0.01 & 0.61 & $+$0.02 & 0.18 & 0.16 & 0.04 \\
902M & 0.46 & 0.59 & $+$0.02 & 0.65 & $+$0.01 & 0.62 & $+$0.01 & 0.25 & 0.16 & 0.04 \\
1.03B & 0.46 & 0.59 & $+$0.02 & 0.65 & $+$0.01 & 0.62 & $+$0.01 & 0.32 & 0.15 & 0.04 \\
1.16B & 0.47 & 0.61 & $+$0.02 & 0.65 & $+$0.01 & 0.62 & $+$0.01 & 0.37 & 0.13 & 0.03 \\
1.29B & 0.47 & 0.62 & $+$0.02 & 0.65 & $+$0.01 & 0.62 & $+$0.01 & 0.39 & 0.11 & 0.03 \\
1.42B & 0.46 & 0.62 & $+$0.02 & 0.64 & $+$0.01 & 0.61 & $+$0.01 & 0.42 & 0.10 & 0.02 \\
1.55B & 0.47 & 0.63 & $+$0.02 & 0.64 & $+$0.01 & 0.61 & $+$0.01 & 0.44 & 0.09 & 0.02 \\
1.68B & 0.47 & 0.62 & $+$0.02 & 0.63 & $+$0.01 & 0.61 & $+$0.01 & 0.47 & 0.08 & 0.02 \\
1.80B & 0.47 & 0.63 & $+$0.03 & 0.64 & $+$0.01 & 0.61 & $+$0.01 & 0.49 & 0.07 & 0.02 \\
1.93B & 0.48 & \textbf{0.65} & $+$0.03 & 0.63 & $+$0.01 & 0.61 & $+$0.01 & \textbf{0.61} & 0.07 & 0.02 \\
\bottomrule
\end{pptable}

\clearpage
\ppappfigure{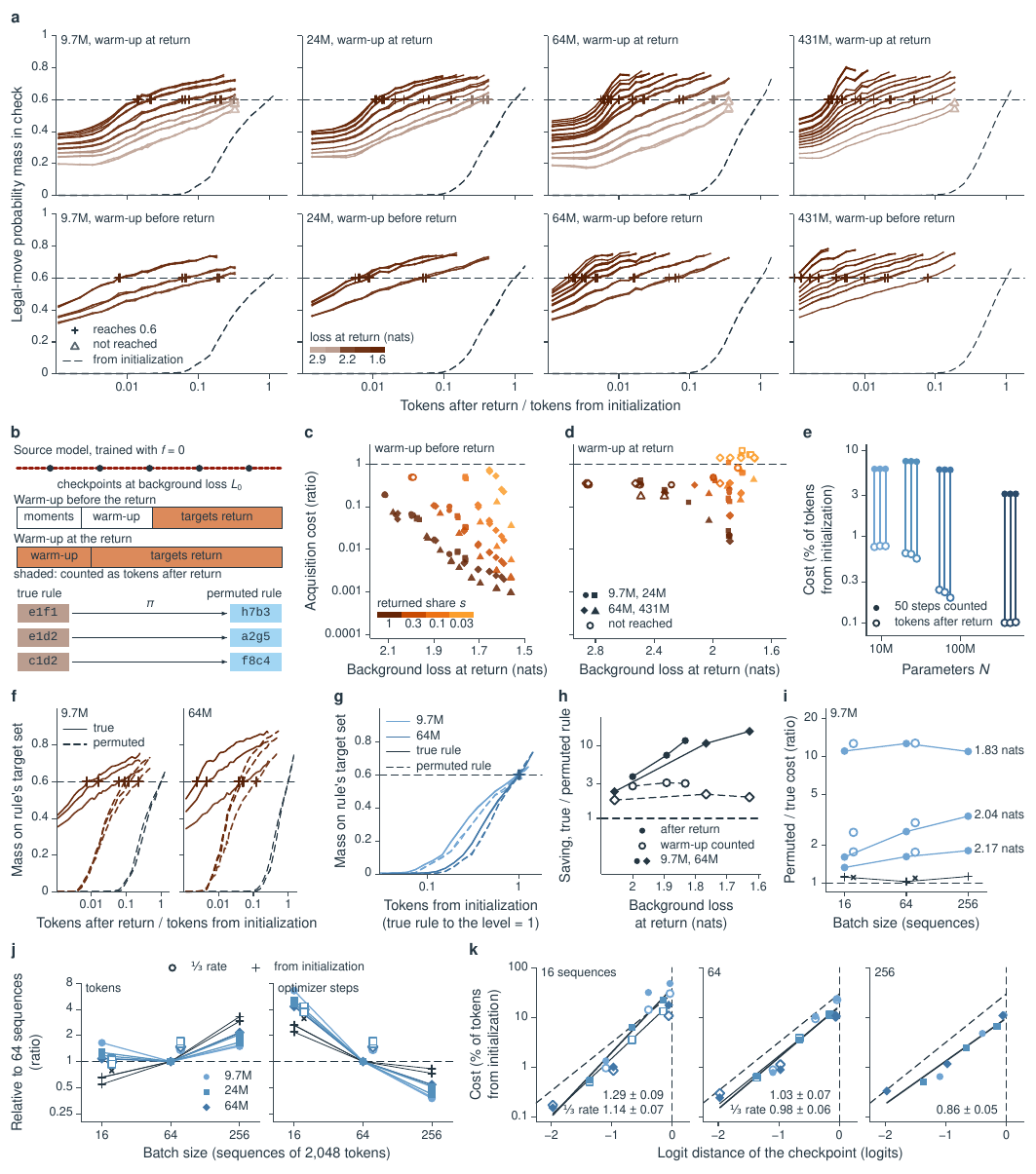}{ed6}{Fig.~D1}{Returned chess targets are acquired at a fraction of their cost, and only the true rule gains.}{Chess source models trained without in-check targets ($f=0$); token axes as in Fig.~3. \textbf{a}, Legal-move mass after all targets return ($s=1$) at every checkpoint of 12 source runs (three per size), warm-up at (top; 89 runs) and before the return (bottom; 52 runs); colour, background loss at return; $+$, 0.6 reached; triangles, not reached. \textbf{b}, Warm-up conditions and permuted rule. \textbf{c},\textbf{d}, Acquisition cost against background loss at return by size (marker) and returned share (colour), warm-up before (\textbf{c}; 110 runs) and at the return (\textbf{d}; 82 runs); open, lower bounds. \textbf{e}, Cost from each source run's lowest-loss checkpoint, without (open) and with (filled) the 50 optimizer-state steps counted. \textbf{f}, Mass on each rule's own target set after the true (solid) or permuted (dashed) rule returns from six checkpoints, each scaled by its own tokens from initialization; colour as in \textbf{a}. \textbf{g}, The two rules from initialization (two seeds per size and rule). \textbf{h}, Ratio of the two savings, true over permuted rule (filled; bars, range over from-initialization seeds), and with the 30 warm-up steps counted for both rules (open). \textbf{i}--\textbf{k}, Returns at other batch sizes and learning rates: three checkpoints each at 9.7M, 24M and 64M, batches of 16, 64 or 256 sequences, full learning rate (filled) or one third of it (open); means of two data orders; plus marks and crosses, from initialization. \textbf{i}, Permuted-rule over true-rule tokens at 9.7M (labels, background loss). \textbf{j}, Tokens and steps relative to 64 sequences at full rate (five checkpoints, returns there of 25--216 steps). \textbf{k}, Cost relative to tokens from initialization at that setting against logit distance (18 runs each); lines, least-squares fits (thin, reduced rate; slopes $\pm$ s.e., decades per logit); dashed, $10^{-0.53+0.97x}$.}

\ppappfigure{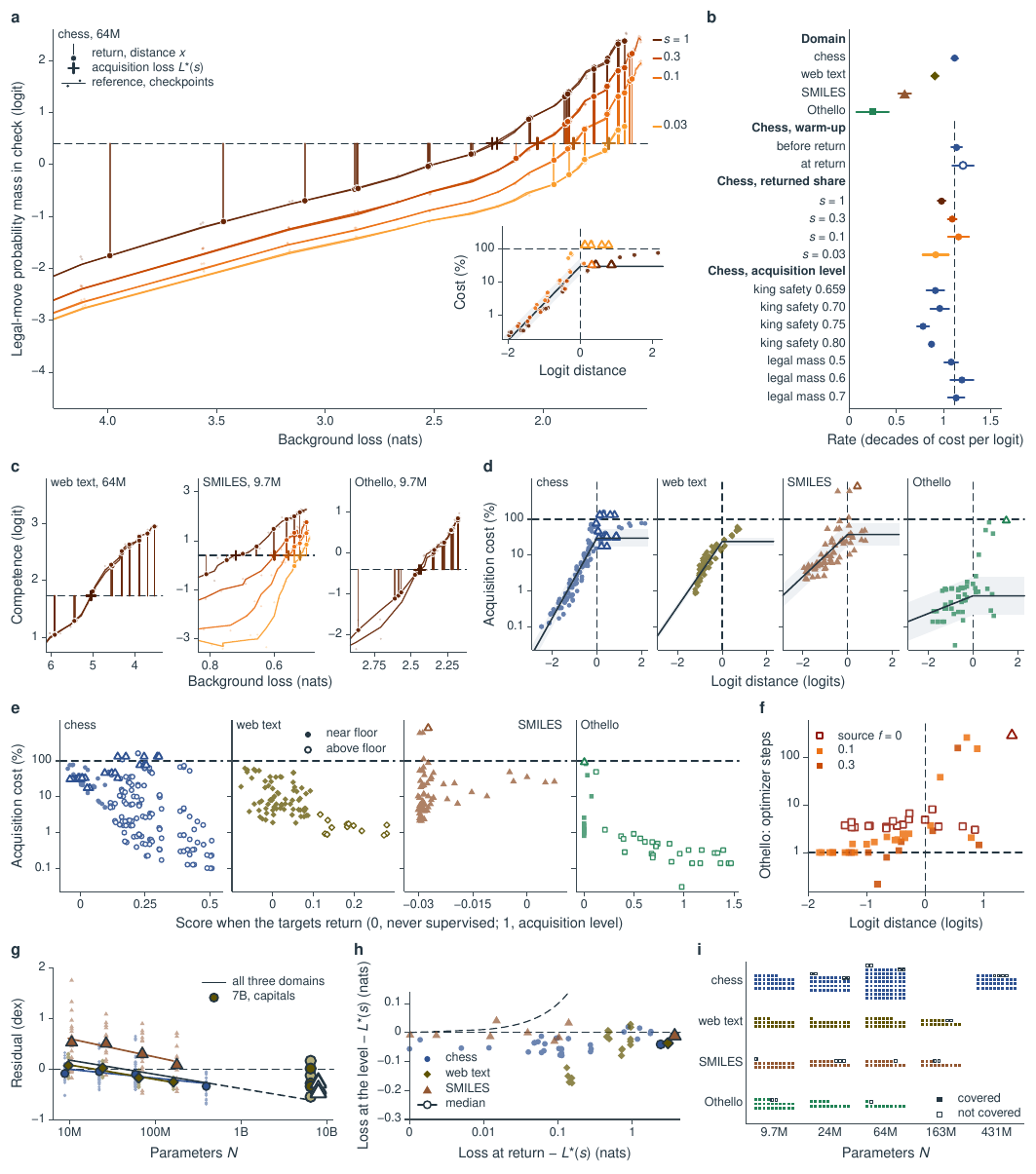}{ed8}{Fig.~D2}{Logit distance orders acquisition cost in every domain, acquisition level and return condition.}{\textbf{a}, Chess 64M: reference trajectories (same-size models trained from initialization with returned share $s$; dots, checkpoints) against background loss; dashed, acquisition level 0.6; crosses, acquisition loss $L^*(s)$; points, 78 return runs (3 source runs) at their return loss; segment, logit distance $x$. Inset, their cost against $x$ with the chess line of Fig.~4a. \textbf{b}, Rate of the two-part relation: per domain from the joint fit of Fig.~4a (Othello, its own fit), and for chess fitted per warm-up condition (before the return, 108 runs; at the return, 51), per returned share (78, 47, 27, 7 runs) and at seven acquisition levels (runs below the acquisition loss); dashed, the chess rate. Bars, $\pm1$ s.e. \textbf{c}, As \textbf{a} for web text (64M, 27 runs), SMILES (9.7M, 20) and Othello (9.7M, 24). \textbf{d}, Cost against $x$ per domain with the domain's line of Fig.~4a (Othello, its own fit) and $\pm1$ residual s.d. (band); triangles, lower bounds. \textbf{e}, Cost against the score at return; filled, the 207 near-floor runs. \textbf{f}, Othello, optimizer steps to the level against $x$ (colour, source target fraction $f$; 50 runs; triangle, not reached). \textbf{g}, Residual from the single least-squares line of Fig.~4b against parameters $N$ (large, size means); lines, fits per domain and pooled (dashed to 7B); large circles and triangles, 7B capitals, measured and lower bounds (dark, seed 0; light, seed 1; not fitted). \textbf{h}, 56 runs returned above the acquisition loss with a positive logit distance: background loss at the crossing minus $L^*(s)$ on a symmetric-log axis; large markers, medians; dashed, crossing at $L^*(s)$ or at return. \textbf{i}, Coverage: one square per run; filled, a reference covers its return loss (390 of 414 runs).}

\clearpage
\section{Returns at 7 billion parameters and of whole domains}\label{app:E}\setcounter{equation}{0}

\subsection{Returned targets at 7B}
Every 7B cost in this subsection and in Table~E1 is relative to the matched fully supervised reference of the same seed. The six snapshots are at steps 2,048, 2,896, 4,096, 5,793, 8,192 and 12,817, and each is continued from the first data block that it has not trained on. The capital targets return with batches of 128 sequences of 2,048 tokens. The five-evaluation smoothing in logit is applied to the returns and to the reference alike.

A sustained crossing requires three consecutive smoothed evaluations at or above the acquisition level. In seed 0 it gives the same cost as the first crossing of the smoothed curve for the returns from steps 12,817 and 8,192, and 22.6\% against 15.2\% from step 5,793; in seed 1 it gives the same cost at all three steps (Fig.~E1c). In seed 0 the first evaluation at or above the level, without smoothing, comes at 0.8\%, 3.0\%, 14\% and 35\% for the last four checkpoints. With only the first answer token supervised, the true assignment reached the level after 1.9\%, 10.8\% and 40.7\% of the reference's tokens from steps 12,817, 8,192 and 4,096 of seed 0. In the 7B model with the capital targets permuted from initialization, the true assignment stayed at or below a competence of 0.08.

Agreement targets, removed in the source model, were returned at the same six checkpoints with an acquisition level of 0.85, which model B reaches after 78.4M tokens in seed 0 and 74.3M in seed 1. The six returns reached it after 2.7--4.0\% of the tokens of their seed's model B in seed 0 and 2.7--5.6\% in seed 1 (Fig.~E1c,e). The source model continued from step 8,192 without returned targets did not reach the capital level (Fig.~E1d), and background loss rose by at most 0.014 nats above its value at the return during a capital return of seed 0 (Fig.~E1f).

\subsection{Domains returned after filtering}
Term accuracy at 7B is read at every evaluation of the returned model and of the continued model B, 162 each. Both streams are unfiltered, so the ratio of tokens is also the ratio of Rust tokens. The filter had left about 259,000 reference-labelled Rust tokens in the 15,000 steps before the return, a realised share of 0.029 and the Rust content of 116.8M tokens of the unfiltered stream at its rate of 0.222\%. With these counted, the Rust tokens of the filtered model when half, three quarters and nine tenths of the gap had closed are 28\%, 13\% and 18\% of those model B had read at the same accuracy. Three quarters of the gap closed after 40M tokens against 1.20B from initialization and nine tenths after 341M against 2.59B. Over the same continuation the gap in Rust loss fell from 0.57 to 0.10 nats, and accuracy on astronomy and Chinese terms stayed within 0.01 of model B's (Fig.~E1i).

Each sampled completion is cut at the brace that closes the function. At the return the filtered model passed the hidden assertions in 3.8\% of completions (95\% interval 2.5--5.1) and model B in 7.3\% (5.0--9.9); after 500M tokens the values were 6.3\% (4.5--8.2) and 7.6\% (5.5--10.0) (Fig.~E1h). The filtered model compiled 19.8\%, 19.9\%, 16.3\%, 16.8\%, 20.0\%, 28.5\% and 37.4\% of its completions after 7.9M, 15.7M, 31.5M, 62.9M, 126M, 252M and 500M tokens. Model B continued on the same blocks compiled 44.7\%, 43.6\% and 41.1\% and passed 7.3\%, 7.5\% and 7.6\% at 0, 126M and 500M tokens.

At 20M--300M the other two domains of each model are filtered as in the Latin square. Shares use five-evaluation smoothing in logit and first interpolated crossings for the returned run and for the unfiltered run from initialization. Table~E2 gives every share. The Chinese model at 300M had closed 42\% of its gap after 5.9\% of the tokens from initialization. The constant was also preferred to a form with a logit-distance term ($-0.15\pm0.38$ per logit) and to one with domain intercepts; the AICc of the three other forms was 3.5--6.0 higher. Within 300 steps, one of the twelve returns closed three quarters of its gap; the others had closed 42--67\%, and their shares exceed 3.8--13.2\%.

\pptablecaption{tab:s1}{Table~E1}{Returned capital targets at 7B.}{Returns of all capital targets to the six checkpoints of the 7B source model in two seeds. The cost is the number of tokens after the return at which smoothed capital competence first reaches 0.20, relative to the tokens at which the matched fully supervised reference reaches it (1.67B in seed 0 and 1.63B in seed 1). The logit distances at step 12,817 are read on model B of seed 0. Returns that did not reach the level within 592M tokens are lower bounds.}
\begin{pptable}{rrrrrr}{0}
\toprule
\pptablehead{Step} & \pptablehead{Seed} & \pptablehead{Background loss} & \pptablehead{Logit} & \pptablehead{Tokens to the} & \pptablehead{Cost (\% of the tokens} \\
 &  & \pptablehead{at return (nats)} & \pptablehead{distance} & \pptablehead{level (millions)} & \pptablehead{of the reference)} \\
\midrule
\endhead
12{,}817 & 0 & 3.002 & $-$0.962 & 29.9 & 1.8 \\
 & 1 & 2.998 & $-$0.917 & 17.6 & 1.1 \\
\addlinespace[2pt]
8{,}192 & 0 & 3.117 & $-$0.454 & 182.6 & 11.0 \\
 & 1 & 3.116 & $-$0.561 & 199.5 & 12.3 \\
\addlinespace[2pt]
5{,}793 & 0 & 3.226 & $+$0.073 & 253.8 & 15.2 \\
 & 1 & 3.223 & $+$0.003 & 362.8 & 22.3 \\
\addlinespace[2pt]
4{,}096 & 0 & 3.341 & $+$0.739 & not reached & $\geq$35.6 \\
 & 1 & 3.340 & $+$0.489 & not reached & $\geq$36.4 \\
\addlinespace[2pt]
2{,}896 & 0 & 3.470 & $+$1.157 & not reached & $\geq$35.6 \\
 & 1 & 3.466 & $+$1.044 & not reached & $\geq$36.4 \\
\addlinespace[2pt]
2{,}048 & 0 & 3.613 & $+$1.597 & not reached & $\geq$35.6 \\
 & 1 & 3.614 & $+$1.678 & not reached & $\geq$36.4 \\
\bottomrule
\end{pptable}

\pptablecaption{tab:s5}{Table~E2}{Domains returned after filtering.}{Models trained with one domain removed (20M--300M, $f=0$) or with Rust filtered to $f=0.03$ (7B), continued on the unfiltered stream. Shares are the tokens after the return at which a fraction of the accuracy gap to the unfiltered model has closed, relative to the tokens at which the unfiltered model reaches the same accuracy from initialization (reference-labelled domain tokens at 20M--300M). A value after $>$ is a lower bound from a return that ended before the fraction was closed. The last column is the fraction of the gap closed at the end of the return, 300 steps at 20M--300M and 0.5B tokens at 7B.}
\begin{pptable}{llrrrrrr}{0}
\toprule
 & & \multicolumn{2}{c}{\pptablehead{Term accuracy}} & \multicolumn{3}{c}{\pptablehead{Share of tokens (\%) to close}} & \\
\cmidrule(lr){3-4}\cmidrule(lr){5-7}
\pptablehead{Domain} & \pptablehead{Size} & \pptablehead{At return} & \pptablehead{Unfiltered} & \pptablehead{Half} & \pptablehead{Three quarters} & \pptablehead{Nine tenths} & \pptablehead{Gap closed at end} \\
\midrule
\endhead
Astronomy & 20M & 0.217 & 0.324 & 3.8 & $>$10.7 &  & 0.65 \\
 & 60M & 0.277 & 0.390 & 7.5 & $>$8.6 &  & 0.58 \\
 & 150M & 0.322 & 0.435 & 3.7 & $>$5.9 &  & 0.64 \\
 & 300M & 0.334 & 0.459 & 1.9 & $>$5.5 &  & 0.62 \\
\addlinespace[2pt]
Rust & 20M & 0.388 & 0.589 & 4.7 & $>$12.4 &  & 0.47 \\
 & 60M & 0.439 & 0.647 & 5.1 & $>$13.2 &  & 0.56 \\
 & 150M & 0.451 & 0.683 & 6.6 & $>$13.2 &  & 0.67 \\
 & 300M & 0.453 & 0.699 & 4.8 & $>$10.4 &  & 0.65 \\
\addlinespace[2pt]
Chinese & 20M & 0.267 & 0.330 & 3.0 & 8.3 &  & 0.80 \\
 & 60M & 0.297 & 0.376 & 6.1 & $>$6.6 &  & 0.58 \\
 & 150M & 0.326 & 0.406 & 6.1 & $>$5.3 &  & 0.53 \\
 & 300M & 0.341 & 0.427 & $>$5.9 & $>$3.8 &  & 0.42 \\
\addlinespace[2pt]
Rust & 7B & 0.517 & 0.764 & 2.1 & 3.4 & 13.2 & 0.95 \\
\bottomrule
\end{pptable}

\clearpage
\ppappfigure{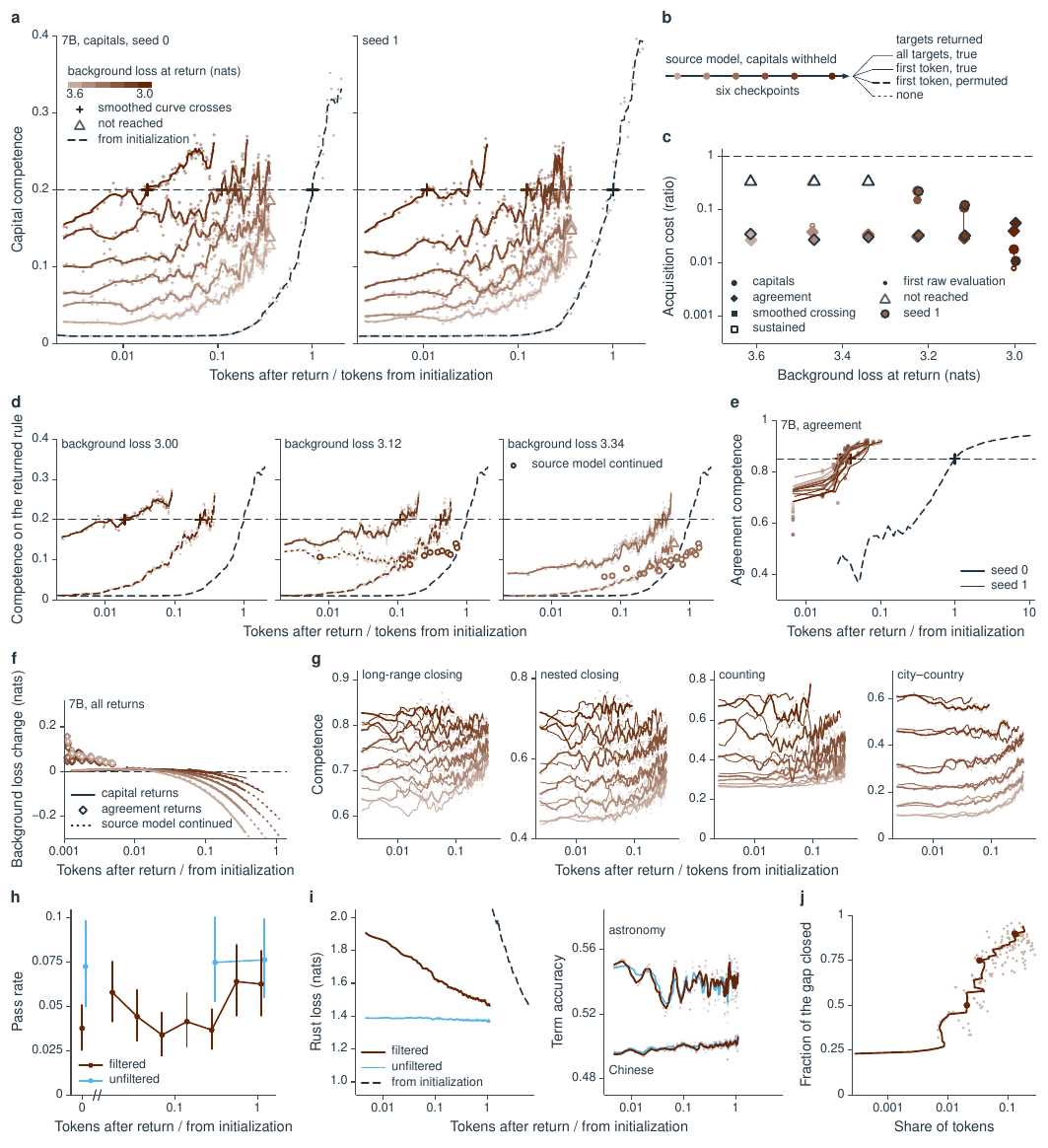}{ed7}{Fig.~E1}{Capitals and Rust return to 7B models at a fraction of their cost.}{Token axes: tokens after the return over the tokens from initialization of the fully supervised model of the same seed to the same acquisition level (dashed curves; dashed horizontal, acquisition level). Lines, five-evaluation smoothing in logit. \textbf{a}, Capital competence after capital targets return at six checkpoints, seeds 0 and 1 (colour, background loss at return; dots, evaluations; $+$, first crossing of 0.20 by the smoothed curve; triangles, not reached). \textbf{b}, Design. \textbf{c}, Acquisition cost of capitals (circles) and agreement (diamonds; level 0.85) against background loss at return: smoothed crossing (filled), sustained crossing (open) and first raw evaluation at the level (small dot); ink-edged, seed 1; triangles, lower bounds. \textbf{d}, Seed 0: all capital targets, and the true and a permuted first answer token returned (line styles as in \textbf{b}); dotted, no targets returned; open circles, the source model's own training. \textbf{e}, Agreement competence after agreement targets return (thin, seed 1); reference, the model of the same seed that keeps agreement targets. \textbf{f}, Background loss minus its value at the return, 31 runs of two seeds (diamonds, agreement returns; dotted, source model continued). \textbf{g}, Competence on skills whose targets were unchanged during the capital returns (thin, seed 1). \textbf{h}--\textbf{j}, A 7B model trained with Rust documents filtered to 3\% after the unfiltered stream returns (brown), and the unfiltered model continued (blue). \textbf{h}, Pass rate of 2,120 sampled completions of 212 function tasks (bars, 95\% bootstrap intervals over tasks; left of the break, at the return). \textbf{i}, Rust loss; astronomy and Chinese term accuracy. \textbf{j}, Fraction of the Rust term-accuracy gap closed against the share of the unfiltered model's tokens from initialization (dots, evaluations; circles, half, three quarters and nine tenths). Unit, one evaluation; Rust, seed 0.}

\ppappfigure{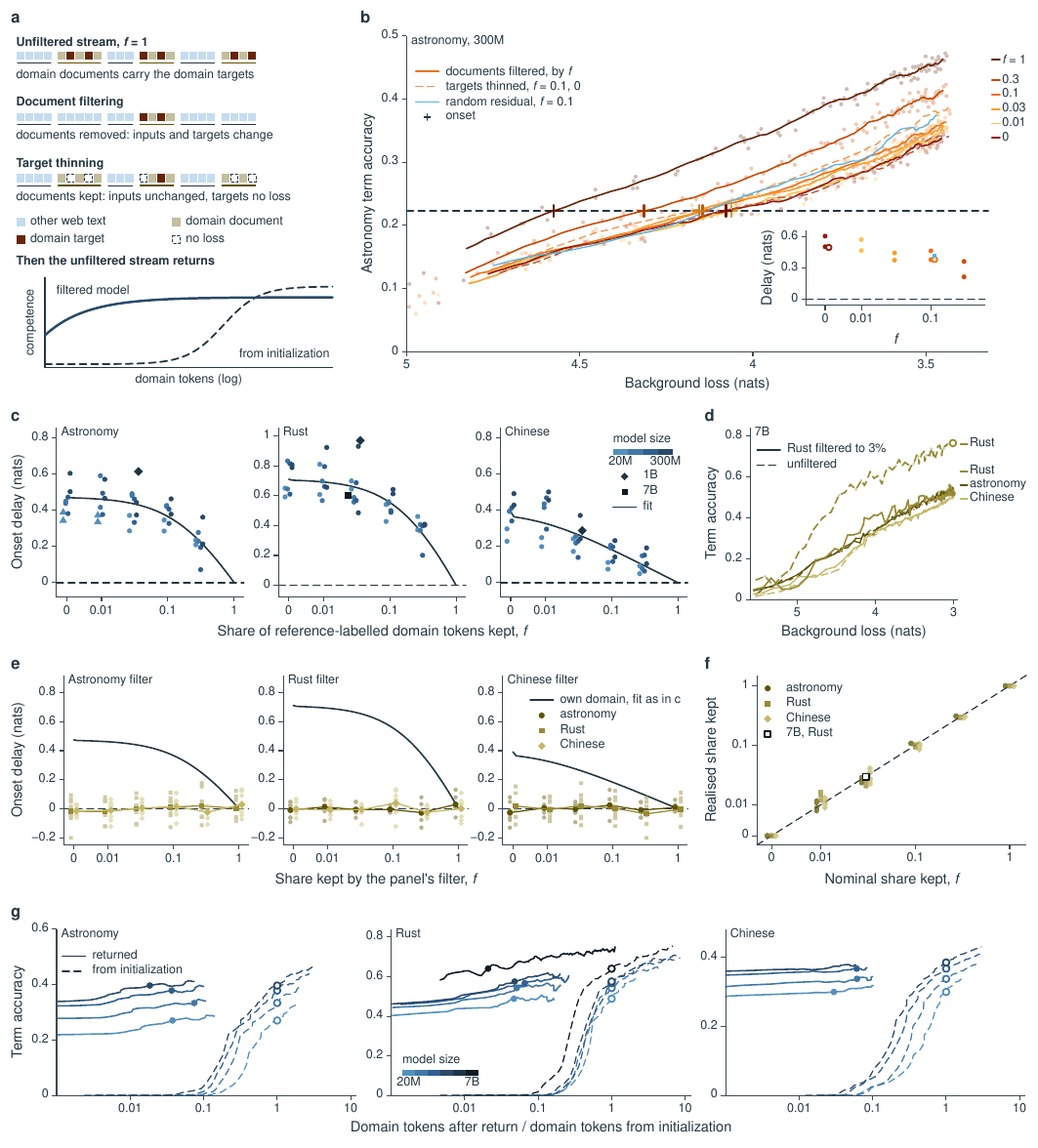}{ed9}{Fig.~E2}{Filtering a domain delays it, and it returns with a few per cent of the domain tokens.}{\textbf{a}, Filtering removes documents with domain tokens; thinning keeps them and removes only the loss on domain targets. \textbf{b}, Astronomy term accuracy against background loss at 300M, filtered to residual share $f=1$ to 0 (dots, evaluations), targets thinned at $f=0.1$ and 0 (dashed) and a random residual at $f=0.1$; $+$, onset. Inset, onset delay against $f$. \textbf{c}, Onset delay relative to the unfiltered model of the same size and seed, 20M--300M (colour; triangles, no onset), 1B at $f=0.03$ and 7B Rust at $f=0.029$; line, fit to 20M--300M. \textbf{d}, Term accuracy of the 7B model with Rust filtered to 3\% (solid) and the unfiltered model (dashed) during training. Circles, the return checkpoint. \textbf{e}, Onset of the other two domains relative to their own fitted onset, against the share kept by each filter (large marks, means); line, the filter's own delay. \textbf{f}, Realised against nominal share kept. \textbf{g}, Term accuracy after the unfiltered stream returns to models trained at $f=0$ (solid; 7B Rust at $f=0.029$) against domain tokens relative to the unfiltered model's at the half-gap level from initialization (dashed); filled circles, half the gap; open, the unfiltered model there. $n$: 48 Latin-square runs (two seeds, c, e, f), 12 at 300M with two thinned and one random-residual run (b); 12 continuation runs (g); one run per model at 1B and 7B, seed 0.}

\clearpage
\section{Removal routes, earlier supervision and restricted returns}\label{app:F}\setcounter{equation}{0}

\subsection{Removal routes}
\textit{Chess.} The 13 cases are 9.7M, 24M, 64M (two seeds) and 163M models at two or three background losses each. Costs relative to initialization use the runs from initialization at 9.7M, 24M and 64M. No 163M model was trained from initialization on the schedule of the return runs, so the two 163M cases are divided by the tokens at which the fully supervised 163M training run of the same seed reaches the level (142.8M). At 9.7M--64M that training schedule reaches the level at 0.59--0.85 times the tokens of the run from initialization on the return schedule, so the 163M ratios are upper estimates relative to the other sizes.

\textit{Web text.} The agreement targets are the six number-marked auxiliaries, and competence is the preference for the auxiliary over its number partner. For each of two seeds, one fully supervised model and one model without agreement targets are trained from initialization for 1.07B tokens; the cases are steps 1,448 and 4,096 of each seed, at background losses of 4.14--4.19 and 3.65--3.67 nats. The withdrawn route continues the fully supervised model from an earlier step without agreement targets until it has read the rows of the never-supervised model at the case step (111M or 315M tokens). Gradient ascent keeps the state of lowest competence whose background loss is at most 0.05 nats above the start. Ascent on the full cross-entropy at the agreement targets lowers the probability of every auxiliary and changes the choice between them little, whereas the two-way loss acts on the choice that competence scores (open and filled diamonds in Fig.~F1i--k). Five unlearned models were at the acquisition level when the targets returned. At the return the background losses of the routes of a case lie within 0.04 nats. In the 30 warm-up steps the learning rate rises linearly to $2.12\times10^{-3}$, evaluations fall at steps 12\% apart beyond step 40, and a return stops after four evaluations at or above 0.85 or at a budget of 700 steps (92M tokens) when all targets return and 1,000 steps (131M tokens) when 10\% return; no return reached its budget. The fully supervised training runs reach the level after 119M and 133M tokens. Never-supervised models reached it with 1.8--2.6\% of those tokens when all targets returned and with 2.3--2.6\% when 10\% returned.

\textit{SMILES.} Competence is the probability of the true closing digit on 8,000 held-out molecules. The withdrawn route continues the fully supervised model from 5 or 12 tokens per parameter without ring-closure targets to the training rows of the case. Gradient ascent uses the cross-entropy at ring-closure targets, which is the loss of the decision that competence scores. Gradient ascent stopped at the limit on background loss and misdirection at a plateau of competence. At the return the background losses of the routes of a case lie within 0.02 nats. Costs are relative to the tokens at which the fully supervised training run of the same size and seed reaches the acquisition level of 0.6. From the same starting competence, never-supervised models needed 5.6--7.9, 5.2--10.7 and 3.7--6.5 times the tokens of withdrawn, gradient-ascent and misdirection models ($n=24$). The removed models reach the level within one to three optimizer steps, so each ratio is resolved to about a factor of 1.5. With all targets returned, never-supervised models reached the level with 2.0--3.0\% of the tokens of the training run. With 10\% returned they reached it in 7 of 8 cases within the budget of 8M tokens, and all 24 other models reached it at 0.27--1.4\%. Without ring-closure targets, models unlearned by misdirection fell to about 0.01 and stood at 0.01--0.09 after 240 further steps, withdrawn models moved by at most 0.05 and never-supervised models by at most 0.003 over the warm-up and the further steps (Fig.~F1k).

\subsection{Earlier supervision}
The reference includes two 9.7M models trained to 1.66 nats with targets returned from 26 checkpoints each, which cover the background loss of every model that kept 3\%. The returns of a compared model are used only inside the reference's range of background loss at its size. The variance between and within source runs of the reference is taken to be the same for the compared models, and the interval uses $t$ quantiles on the reference's degrees of freedom.

\subsection{Related king-safety positions}
Inputs stay intact in every family-masking run, and background loss is measured on targets outside check and outside the masked families. The nine background losses are equally spaced. In the range over the 16 assignments of a seed to each fraction, each assignment is refitted with its own $\kappa(L)$ and $\tau(L)$.

\subsection{Restricted returns, representation edit and stitched models}
\textit{Restricted returns.} A trainable block comprises both norms, the attention and the MLP; every block is tested in one source run and the first, middle and last block in the other. Adapters $Wx+BAx$ are placed on all seven linear maps of every block, with $A$ uniform in $\pm d_{\mathrm{in}}^{-1/2}$ and $B=0$ at the start, and all other weights are held fixed. Trainable groups keep their learning rates, and adapters take the rate of the hidden matrices without weight decay. Where no multiplier gave a permuted return that reached the level, the multiplier with the highest final permuted mass was used; the unrestricted return uses the standard learning rate. The unrestricted returns cost 0.23--0.78\% of the tokens from initialization and the permuted rule 12.0--16.6 times more. A single block in the upper half cost 4.1--7.0, 2.6--4.2 and 2.2--2.8 times the unrestricted return at 9.7M, 24M and 64M. The first block alone cost 39--50 times at 64M and did not reach the level at 9.7M and 24M, and the unembedding with the final norm cost 12--13 times at 64M and did not reach it at 9.7M and 24M. Adapters of rank 1, 4 and 16 cost 3.3--58, 2.2--7.4 and 1.3--3.1 times. Both rules reached 0.6 under 19 restrictions, only the true rule under 36 and neither under 9. The best multiplier lay at the top of the grid for rank-1 adapters, for rank-4 adapters at 64M and for the rank-16 adapter of one 64M run, so the adapter costs are upper estimates. The experiment comprises 396 runs.

\textit{Representation edit.} The three checkpoints of each source run are at 10, 20 and 64 tokens per parameter at 9.7M and at 5, 12 and 16 at 24M. The counterfactual position pairs come from the 30\% of games held out from the fit of the subspace, and the edit replaces the subspace by its mean for the side to move. A checkpoint was used if the edit removed at least half of the readout, raised background loss by at most 0.20 nats and removed at least five times as much as random rank-8 ablations; all 24 checkpoints met these conditions.

\textit{Stitched models.} At 9.7M (8 blocks, two seeds) the late checkpoint is at 619M tokens and the early ones at 290M and 193M tokens, where the unrestricted return costs 3.2--3.6 and 7.9--8.4 times as much, with $K=4$ in both seeds and $K=2$ and 6 in one. At 64M (13 blocks, one seed) the checkpoints are at 644M and 161M tokens, a 6.7-fold difference in cost, with $K=3$, 6 and 10. Every stitched model continues on the data that follow the late checkpoint. The two data orders agreed within 23\%, and every run reached 0.6. After the 30 warm-up steps the background loss of each stitched model lay within 0.01 nats of the range spanned by its two checkpoints. The intact models that define the line are three at 9.7M and two at 64M. At 9.7M the position of a stitched model between its two checkpoints followed the number of late blocks: 0.36--0.43, 0.55--0.60 and 0.70--0.87 of the range in log cost with six, four and two late blocks. At 64M late upper blocks lowered background loss, and with it the cost, more than late lower blocks (0.28 against 0.65 of the range at $K=6$). The ratio of permuted to true cost followed the true-rule cost: 16.3 and 4.3 for the intact 64M checkpoints, 11.8 and 6.0 for the stitched models at $K=6$. The experiment comprises 96 runs.

\clearpage
\ppappfigure{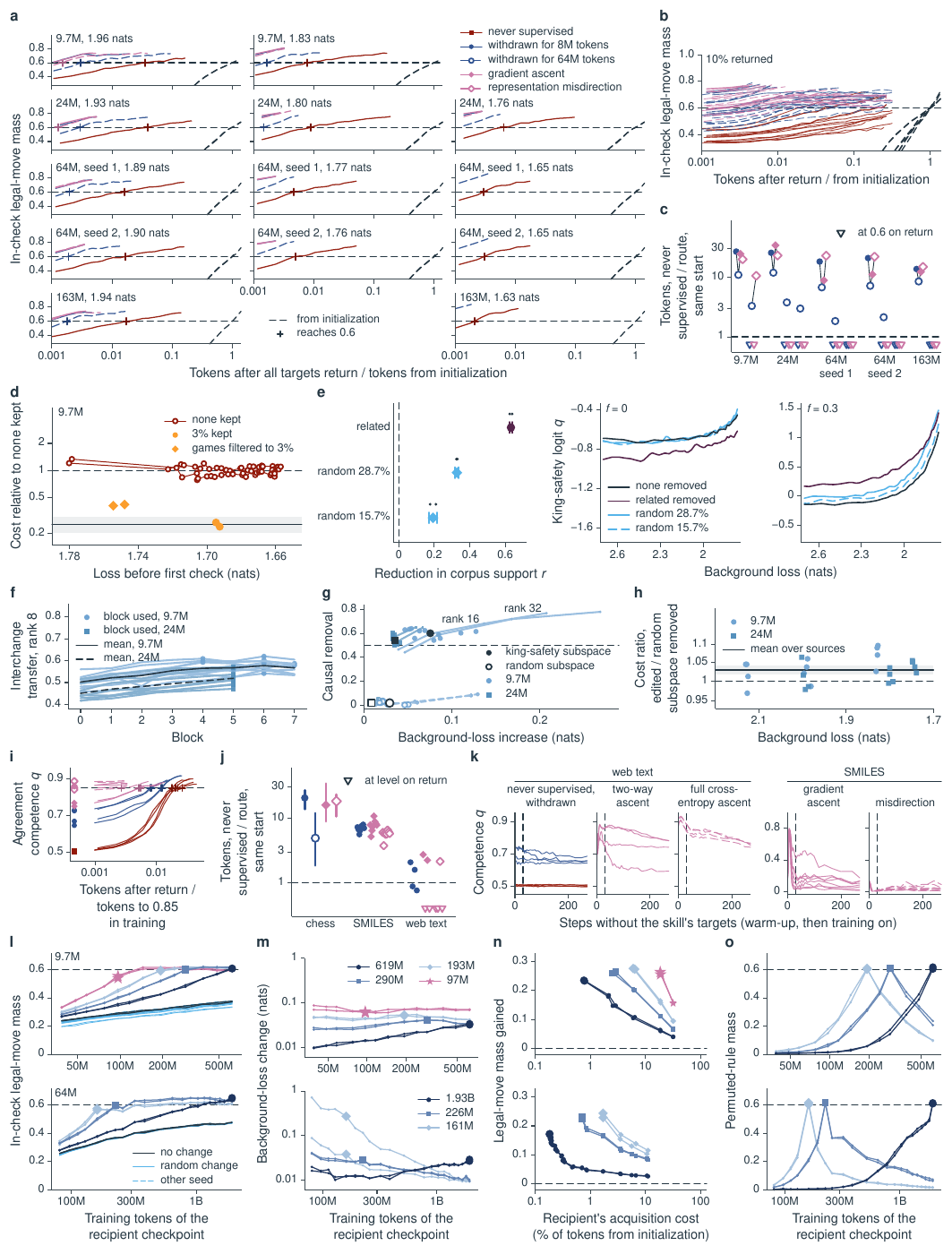}{ed10}{Fig.~F1}{Cost of return by removal route, and transplanted weight changes.}{\textbf{a}--\textbf{c}, Chess, 13 cases (9.7M--163M), never-supervised models and four removal routes (key in \textbf{a}). \textbf{a}, Legal-move mass after all targets return (tags, size and background loss; tokens as in Fig.~5a; crosses, 0.6 reached). \textbf{b}, As \textbf{a} for 10\% returned (65 models). \textbf{c}, The ratio of Fig.~5b by case ($n=26$; lines join a case); triangles, at 0.6 on return ($n=26$). \textbf{d}, As Fig.~5e. \textbf{e}, As Fig.~5f. \textbf{f}--\textbf{h}, Representation edit, 24 checkpoints of eight 9.7M and 24M runs: interchange transfer of a rank-8 subspace by block (\textbf{f}; filled, block edited); removed influence of attacked squares on king moves against background-loss increase, king-safety and random subspaces (\textbf{g}; thin, ranks 4--32); tokens to 0.6 relative to a random subspace (\textbf{h}). \textbf{i}--\textbf{k}, Removal routes in web text (300M, subject--verb agreement; four cases) and SMILES (ring closure, 9.7M and 24M; eight cases); markers as in \textbf{a} (diamonds: web-text ascent on the two-way loss, filled, or full cross-entropy, open). \textbf{i}, Web-text competence after all targets return; crosses, 0.85 reached. \textbf{j}, As \textbf{c} for chess (route means and ranges), SMILES ($n=24$) and web text ($n=7$; triangles, at the level on return, $n=5$). \textbf{k}, Competence during the warm-up and 240 steps without the skill's targets. \textbf{l}--\textbf{o}, Transplanted weight changes, chess, two source runs each at 9.7M (top) and 64M (bottom): the change due to returned targets at one checkpoint (key in \textbf{m}, training tokens; large markers), added without training to other checkpoints of its run (13 and 22). \textbf{l}, Legal-move mass; ink, no change; control colour, equal-norm Gaussian change (solid) or other-seed recipients (dashed). \textbf{m}, Background-loss change. \textbf{n}, Mass gained against the recipient's acquisition cost, checkpoints no later than the change's. \textbf{o}, As \textbf{l} for a permuted-rule return, read on that rule.}

\clearpage
\section{Statistics}\label{app:G}\setcounter{equation}{0}

In the analyses of acquisition cost a point is one return run, and a source run is one training run at one seed, shared by its checkpoints. Least-squares rates for single domains are given with the larger of the ordinary and the cluster-robust standard error. Along a thinned trajectory, the offsets at interpolated background losses are repeated measurements of that trajectory, so errors of the derived relation are taken over models and held-out prediction errors are given per trajectory. Probabilities of output classes have intervals from 1,000 bootstrap resamples of games, and the isolated web-text contrasts have $t$ intervals over four seeds. The family-masking contrast has two paired seeds, so the range over the 16 recombinations of seeds is shown in place of a sampling interval.

\end{document}